%% file: main.tex
\documentclass{article}
\PassOptionsToPackage{table}{xcolor} 

\usepackage[preprint]{corl_2026} %

\usepackage{xcolor}
\usepackage{colortbl}
\usepackage{makecell}   
\usepackage{amsmath}
\usepackage{graphicx}
\usepackage{multirow}  
\usepackage{array}
\usepackage{tabularx}
\usepackage{wrapfig}
\usepackage{caption}
\usepackage{booktabs}

\usepackage{adjustbox}
\usepackage{tcolorbox}
\tcbuselibrary{breakable, listings}
\usepackage{listings}
\usepackage{enumitem}

\usepackage[table]{xcolor}
\definecolor{helmtint}{HTML}{EFF8E6}

\usepackage{hyperref}

\title{%
  HeLM: Hierarchical Explicit Language Memory\\
  for Non-Markovian Manipulation
}

\author{%
  Gyeonghun Kim \quad
  Donghoon Kim \quad
  Minsoo Choi \quad
  Suyun Lee \\[4pt]
  Seoul National University \\[3pt]
  {\small\url{https://kimspacehuni.github.io/HeLM/}}
}

\begin{document}

\hypersetup{
  pdftitle={HeLM: Hierarchical Explicit Language Memory for Non-Markovian Manipulation},
  pdfauthor={Gyeonghun Kim, Donghoon Kim, Minsoo Choi, Suyun Lee},
  pdfsubject={Robotic manipulation with explicit language memory},
  pdfkeywords={Robot Learning, Vision-Language-Action Models, Language Memory, Non-Markovian Manipulation}
}

\raggedbottom

\maketitle

\begingroup
  \renewcommand{\thefootnote}{}
  \footnotetext{%
    Correspondence to:
    \href{mailto:kgh011016@snu.ac.kr}
         {\nolinkurl{kgh011016@snu.ac.kr}}.\\
    Other authors:
    \texttt{\{dhkim, mschoi, sylee\}@islab.snu.ac.kr}.
  }
\endgroup

\input{section/abstract}
\input{section/introduction}
\input{section/related_work}
\input{section/method}

\input{section/experiments}
\input{section/conclusion}

\bibliography{references}

\appendix
\clearpage
This supplementary material presents additional implementation details and experimental results for our paper,
\emph{HeLM: Hierarchical Explicit Language Memory for Non-Markovian Manipulation}.
The appendix provides extended implementation details, detailed task descriptions, failure analyses, and supplementary experiments that further support the conclusions of the main paper.
Additionally, we include a demo video illustrating the overall framework and real-world inference examples of HeLM.
\vspace{0.1cm}

\input{section/appendix} 

\end{document}

%% file: section/abstract.tex
\begin{abstract}
Recent advances in Vision-Language-Action (VLA) models have enabled fine manipulation capabilities, yet performing tasks that require working memory remains a key challenge.
Most existing VLA architectures operate under a Markovian assumption, mapping only the current observation to an action, which leads to failure in non-Markovian environments.
Inspired by the human `inner monologue', we propose Hierarchical Explicit Language Memory (HeLM), a hierarchical VLA framework that maintains task-relevant history in explicit language through factorized event detection and event-triggered memory updates.
We evaluate HeLM across 10 non-Markovian real-world tasks with 56 variants, including five newly formulated inter-episode tasks that require persistent context across task boundaries.
HeLM improves intra-episode success by 12.8 percentage points over the strongest baseline and achieves 78\% subtask success on compositional generalization tasks, while using up to 4.5$\times$ less memory and running 2.5$\times$ faster than the keyframe baseline at long horizons.

\end{abstract}

%% file: section/introduction.tex
\section{Introduction}

The rise of Vision-Language-Action (VLA) models has revolutionized robotic control by enabling generalization across diverse language instructions and embodiments~\cite{kim2024openvlaopensourcevisionlanguageactionmodel, pi0}.
Built on large-scale pretrained vision-language models, VLA models generate robot actions based on visual observations and natural language instructions.
This approach has led to widespread deployment, with variants focusing on visual grounding~\cite{zheng2025tracevlavisualtraceprompting, li2025fastmemorybaseddataefficientvisionlanguage}, complex reasoning capabilities~\cite{zawalski2025roboticcontrolembodiedchainofthought, huang2025thinkactvisionlanguageactionreasoningreinforced, huang2022innermonologueembodiedreasoning}, or subtask planning~\cite{shi2025hirobotopenendedinstruction}.

While recent advancements in robot policies have enhanced fine manipulation capabilities, tasks requiring working memory remain a key challenge for real-world deployment.
When wiping the window, the robot cannot determine whether the surface has already been wiped and ends up wiping the same area repeatedly.
Such tasks are fundamentally non-Markovian, meaning that the current observation alone is insufficient to determine the optimal action.
Yet most existing VLA architectures operate under a Markovian assumption, treating manipulation as a sequence of independent decisions.
These structural limitations cause failure in memory-intensive tasks (see Fig. ~\ref{fig:motivation}).

\input{figure_tex/figure_1}
Recently, various approaches have been proposed to address such limitations~\cite{pomdpkaelbling1998planning}.
One line of work extracts latent embeddings from historical observations~\cite{liberomem, fang2025sam2actintegratingvisualfoundation} or compresses tokens into a memory module~\cite{memoryvla}.
While these approaches are effective to some extent, relying on latent representations makes it difficult to identify causes of failure due to limited interpretability.
Another line of work directly leverages visual history by stacking past frames~\cite{torne2025learninglongcontextdiffusionpolicies} or using keyframe annotations~\cite{memer}.
A drawback of these approaches is computational overhead caused by accumulating image tokens, which limits long-horizon scalability.
\vspace{1cm}

To overcome these limitations, we draw inspiration from how humans utilize an `inner monologue'.
Just as humans manage complex tasks by internally keeping track of their progress (e.g., ``I pressed the button twice; one more to go") or spatial data (e.g., ``The cup was originally on the left"), we suggest that natural language can serve as an efficient and interpretable representation for robotic memory.
By nature, this compact form preserves critical information (e.g., counts, locations) across long-horizon tasks.
More fundamentally, language is a natural memory representation for VLM-based policies: it is the native modality of vision-language reasoning, scales compactly across long horizons, and exposes inspectable memory states for analysis.

Building on this insight, we propose Hierarchical Explicit Language Memory (HeLM), which incorporates language memory into robotic manipulation.
HeLM separates high-level memory management from low-level action generation.
Specifically, a VLM-based high-level policy (HLP) maintains the task relevant history in language form and generates action commands, while a VLA-based low-level policy (LLP) executes actions.
Memory updates are event-triggered and occur only after task-relevant events or instruction changes.
The language memory is organized into working and episodic fields according to whether information is needed within or across instructions.
Together, these design choices support non-Markovian reasoning both within and across episodes.

We evaluate HeLM on diverse non-Markovian tasks, including newly formulated inter-episode tasks.
HeLM improves intra-episode subtask success by 12.8 percentage points over the strongest baseline and achieves 78\% on compositional HLP generalization tasks.
As task length grows, HeLM uses up to 4.5$\times$ less memory and runs 2.5$\times$ faster than the visual-memory baseline at long horizons.

\textbf{Contributions.} Our main contributions are summarized as follows:
\begin{itemize}[topsep=2pt, itemsep=2pt, parsep=0pt, leftmargin=*]
\item We propose HeLM, a hierarchical framework that maintains task-relevant history in explicit language through factorized event detection and event-triggered memory updates.
    
\item Across 10 real-world tasks, HeLM outperforms memory-augmented and visual memory baselines, supported by component-level failure analysis.
    
\item We formulate inter-episode persistence as an underexplored axis of robotic memory and introduce a task suite of 10 non-Markovian tasks (56 variants). %
\end{itemize}

%% file: figure_tex/figure_1.tex
\begin{figure}[!t]
    \centering
    \includegraphics[width=\linewidth]{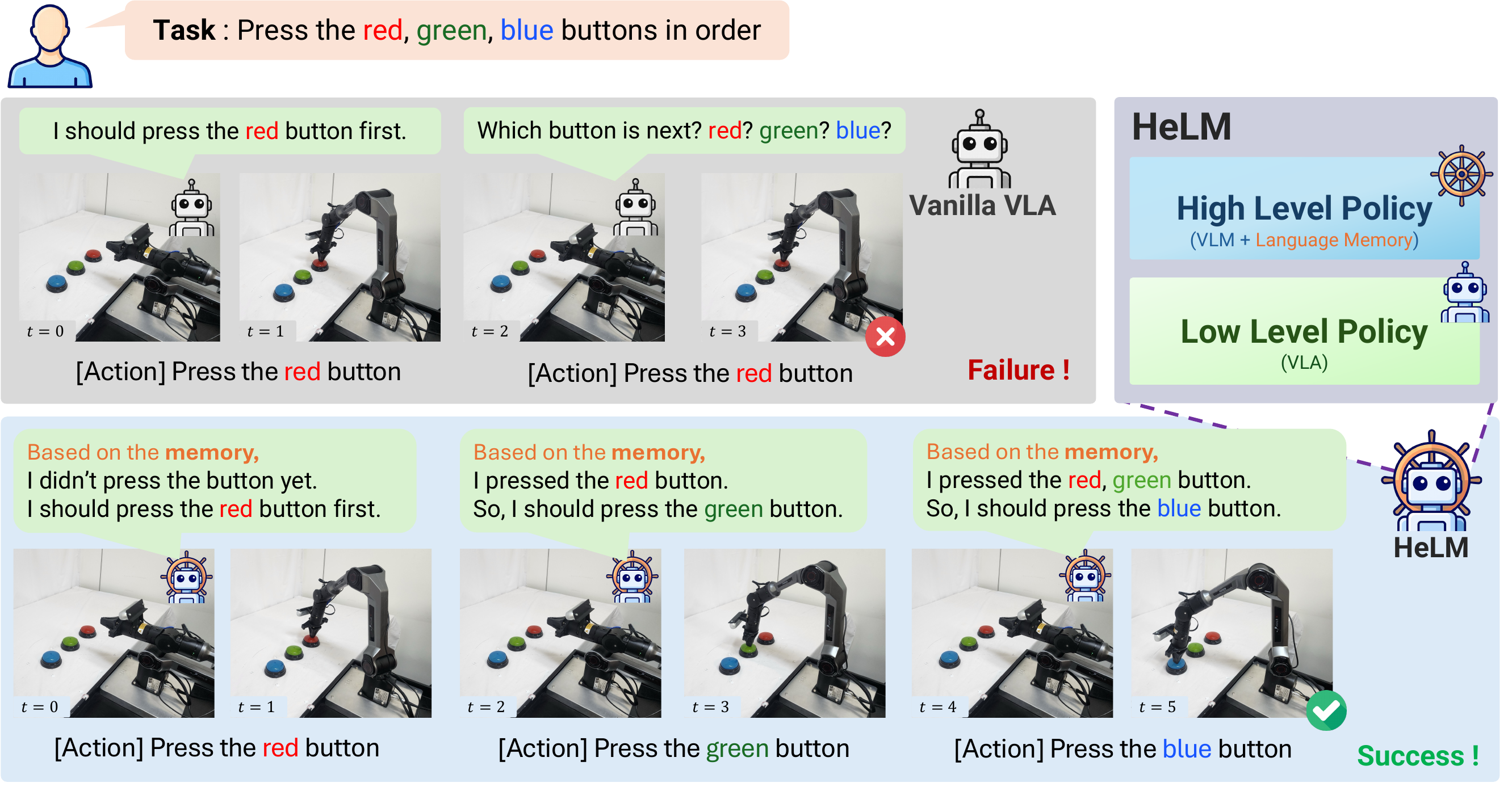} 
    \caption{\textbf{Example of non-Markovian task.} Due to architectural limitation, a vanilla VLA loses track after pressing the red button and presses it again. HeLM successfully executes the task by storing progress in language memory, allowing the HLP to guide the LLP correctly.
    }
    \label{fig:motivation}
\end{figure}

%% file: section/related_work.tex
\section{Related Work}
\label{sec:related_work}

\subsection{Memory-Augmented and Hierarchical VLA Models}
VLA models such as RT-2, OpenVLA, and $\pi_0$ map current visual-language observations to robot actions~\cite{brohan2023rt2visionlanguageactionmodelstransfer,kim2024openvlaopensourcevisionlanguageactionmodel,pi0,kim2026rldx1technicalreport}.
Recent works extend these models with richer reasoning capabilities ~\cite{geminiroboticsteam2025geminiroboticsbringingai,intelligence2025pi05visionlanguageactionmodelopenworld,onetwovla,foreact}, while hierarchical VLA frameworks decompose complex tasks into high-level instructions executed by low-level policies~\cite{shi2025hirobotopenendedinstruction}.
However, these approaches do not necessarily maintain explicit task-relevant history over long horizons.

Prior memory-augmented policies retrieve latent features or historical tokens~\cite{memoryvla,mapvla,lin2025echovlaroboticvisionlanguageactionmodel,pragmabot,liberomem}, while others preserve visual history through feature traces or multi-frame observations~\cite{zheng2025tracevlavisualtraceprompting,chen2025historyawarevisuomotorpolicylearning,hu2025resolvingstateambiguityrobot,jang2025contextvlavisionlanguageactionmodelamortized,memer}.
These representations can retain historical information, but provide limited access to the explicit task state required for counting, sequencing, or cross-episode recall.

Language has also been used to support temporal reasoning.
Inner Monologue~\cite{huang2022innermonologueembodiedreasoning} provides language feedback such as success signals and scene descriptions to support closed-loop replanning, whereas HeLM explicitly keeps task-relevant history in language over the sequence.
MEM~\cite{mem} combines short-horizon visual context with long-horizon textual memory, whereas HeLM separates event detection from memory updates, updating language memory only when task-relevant events are detected or instructions change.
HeLM further evaluates inter-episode tasks, where completing a later instruction requires information acquired during an earlier episode.
RoboMemory~\cite{robomemory} stores and retrieves past experience to support future interactions, whereas HeLM focuses on maintaining task-relevant history for the current non-Markovian task.
Chameleon~\cite{chameleon} also uses a two-part memory but operates over latent multimodal tokens rather than explicit language.
Overall, HeLM studies explicit language memory as a compact and interpretable way to carry task-relevant information over time and across instructions.

\subsection{Non-Markovian Manipulation Tasks in Robotics}
Manipulation benchmarks such as LIBERO and RLBench evaluate episodes independently, with environment resets and no requirement to retain information across task boundaries~\cite{rlbench, liu2023liberobenchmarkingknowledgetransfer, robocasa, mikasarobo, li2024evaluatingrealworldrobotmanipulation}.
Recent memory-focused benchmarks such as MemoryBench, LIBERO-Mem, RMBench, and RoboMME~\cite{fang2025sam2actintegratingvisualfoundation, liberomem, rmbench, robomme} introduce historical dependencies such as object occlusion, action history, and sequential progress.
However, these benchmarks primarily evaluate memory within individual task episodes.
In contrast, our task suite includes inter-episode tasks, where completing a later instruction requires information acquired during an earlier episode that cannot be recovered from the current observation alone.

%% file: section/method.tex
\input{figure_tex/figure_2}

\section{HeLM: Hierarchical Explicit Language Memory}
\label{sec:method}

HeLM's design follows a single principle: language can serve as a memory representation for VLA-based manipulation. 
This section formalizes the non-Markovian problem (Sec.~\ref{subsec:problem_formulation}), hierarchical decomposition (Sec.~\ref{subsec:overview}), structured memory (Sec.~\ref{subsec:structured_memory}), and event-triggered inference (Sec.~\ref{subsec:two_phase}).

\subsection{Problem Formulation}
\label{subsec:problem_formulation}
In non-Markovian manipulation tasks, the current observation $o_t$ is insufficient to fully infer the true state $s_t$.
Standard VLA policies model the action distribution as $\Pi(a_t \mid o_t, I)$, implicitly assuming that $o_t$ provides sufficient state information.
However, identical observations can correspond to different underlying states depending on past events:
\begin{equation}
P(s_t \mid o_t, I) \neq P(s_t \mid o_{0:t}, I). 
\label{eq:non_markovian}
\end{equation}
A memoryless policy conditioned only on $o_t$ cannot distinguish histories that yield the same current observation but require different actions (see Fig.~\ref{fig:motivation}).

To address this gap, we introduce an explicit, structured textual memory $M_t$ that captures task-relevant history, reformulating the policy as $\Pi(a_t \mid o_t, I, M_t)$.

\subsection{Framework Overview}
\label{subsec:overview}
HeLM decouples semantic memory management from action generation through a two-level hierarchy.
As shown in Fig.~\ref{fig:overall_architecture}, the HLP ($\pi_{HLP}$), built on a fine-tuned VLM, generates action commands $z_t$ while maintaining the memory $M_t$ based on $o_t$ and $I$.
On the other hand, the LLP ($\pi_{LLP}$), implemented with a VLA model, maps HLP-generated commands $z_t$ and $o_t$ to robot actions $a_t$.
By design, language memory captures abstract task state, while the current observation provides immediate spatial grounding at execution time.

\subsection{Structured Memory $M_t$}
\label{subsec:structured_memory}
Inspired by the distinction between working and episodic memory in human cognition, we organize textual memory into two fields based on how long information remains relevant (Fig~\ref{fig:overall_architecture}):
\begin{equation}
    M_t = \langle m^{\text{work}}_t, m^{\text{epi}}_t \rangle.
\end{equation}

Here, working memory $m^{\text{work}}_t$ tracks active progress within a single episode and is updated whenever a visual event is detected.
Episodic context $m^{\text{epi}}_t$ preserves summarized information across task boundaries, which is promoted from $m^{\text{work}}_t$ at the end of each episode (see Fig. 3 for examples).
\input{figure_tex/figure_3}

\subsection{Two Stage HLP Inference}
\label{subsec:two_phase}
HeLM employs a two-stage inference mechanism that updates memory only when semantic events are detected or the instruction changes.
The LLP executes an action chunk of horizon $H$, and the HLP is synchronously invoked at each chunk boundary for event detection and updates memory when needed.
This decoupling avoids unnecessary memory updates between detected events.
As shown in Fig.~\ref{fig:overall_architecture}, $\pi_{\text{det}}, \pi_{\text{upd}}$ share the same VLM parameters and are functionally distinguished through different system prompts.

\vspace{0.1cm}
\noindent\textbf{Visual Event Detection Stage. ($\pi_{\text{det}}$)}:
In Detect Mode, the VLM determines whether a task-relevant event has occurred from the current observation $o_t$, memory $M_t$, and action command $z_t$.
The detector outputs either a task-relevant event or \texttt{none} at each HLP invocation.
\begin{equation}
    e_t \sim \pi_{\text{det}}(e_t \mid o_t, M_t, z_{t}).
\end{equation}
These events are defined via a predefined set of task-relevant events, specified via TaskSpec (Sec.~\ref{subsec:training}) such as \texttt{"pressed red button"} or \texttt{"wiped the top window"}.

\vspace{0.1cm}
\noindent\textbf{Memory Update Stage. ($\pi_{\text{upd}}$)}:
In Update Mode, the HLP updates the language memory based on the detected event and current memory state.
This phase is event-triggered and is activated only when the detector identifies an event (i.e., $e_t \neq \emptyset$) or when the global instruction changes ($I_t \neq I_{t-1}$).
\begin{equation}
    M_{t+1}, z_{t+1} = 
    \begin{cases} 
        \pi_{\text{upd}}(I_t, M_{t}, e_t) & \text{if } e_t \neq \emptyset \text{ or } I_t \neq I_{t-1} \\
        M_{t}, z_{t} & \text{otherwise}
    \end{cases}
\end{equation}
\subsection{Data Generation}
\label{subsec:training}
HeLM's explicit text memory formulates training as standard visual instruction tuning, where each sample pairs a visual observation with a text prompt encoding the current memory state.
Training data is synthesized from short subtask episodes via TaskSpec, a logical schema that can be hand-written or LLM-generated.
TaskSpec is used to synthesize DETECT and UPDATE supervision from short subtask trajectories.
TaskSpec specifies the valid action commands, task-relevant events, and corresponding memory updates for each task, without prescribing a fixed action sequence.
During inference, the HLP receives task-specific event and action vocabularies, but not the full TaskSpec or ground-truth memory transitions.
Instead, it autonomously detects events and updates its language memory based on visual observations.
Appendix~\ref{app:helm_implementation} details the data-generation pipeline, TaskSpec schema, system prompts, and training hyperparameters.

\section{Non-Markovian Task Suite}
\label{subsec:task}
To evaluate non-Markovian reasoning, we formalize inter-episode setting and introduce a task suite covering both intra- and inter-episode challenges.

\subsection{Extension to Inter-Episode Setting}
\label{subsec:inter_episode}
Extending the formulation in Section~\ref{subsec:problem_formulation} (fixed instruction $I$), we now consider settings where the agent receives a sequence of instructions $I_1, I_2, ...$, with episode boundaries marked by instruction changes.
Each episode follows a single task instruction.
At an episode boundary, the robot returns to its initial pose and the within-episode history is cleared, while the episodic context is retained.
We define \textit{inter-episode tasks} as those where completing a later instruction requires information acquired during an earlier episode that cannot be recovered from the current observation alone.

For an episode starting at $t_k$, within-episode history $o_{t_k:t}$ alone may be insufficient:
\begin{equation}
    P(s_t \mid o_{t_k:t}, I_t) \neq P(s_t \mid o_{0:t}, I_{0:t}).
    \label{eq:inter_episode}
\end{equation}
Eq.~(\ref{eq:non_markovian}) is the single-episode case of this.
HeLM handles this through $m^{\text{epi}}_t$, which persists across episode boundaries while $m^{\text{work}}_t$ resets.
At each boundary, $\pi_{\text{upd}}$ (triggered by $I_t \neq I_{t-1}$) promotes summarized history from $m^{\text{work}}_t$ into $m^{\text{epi}}_t$.

\subsection{Task Suite Design}
\label{subsec:task_design}
Our task suite consists of 10 tasks with 56 variants and covers three cognitive challenges (Fig.~\ref{fig:experiment_task}):
count-dependent reasoning, order-sensitive progress tracking, and object-centric recall after occlusion or displacement.

We further categorize tasks by temporal dependency.
Intra-episode tasks (A1--A5) require tracking information within a single instruction, while inter-episode tasks (E1--E5) require information from an earlier instruction to complete a later one.
The intra-episode tasks involve remembering how many times an action has been repeated (A1), which actions have already been completed in a sequence (A2), which regions of the window have been wiped (A3), which drawers have already been searched (A4), and the current stage of a multi-step manipulation (A5).
The inter-episode tasks require recalling a previous count (E1), a human-demonstrated action sequence (E2), previously wiped regions (E3), an observed object location (E4), or an object's original position (E5).
Instruction templates, task variants, and subtask decompositions are provided in Appendix~\ref{app:task_detail}.

%% file: figure_tex/figure_2.tex
\begin{figure*}[t] %
    \centering
    \includegraphics[width=1\textwidth]{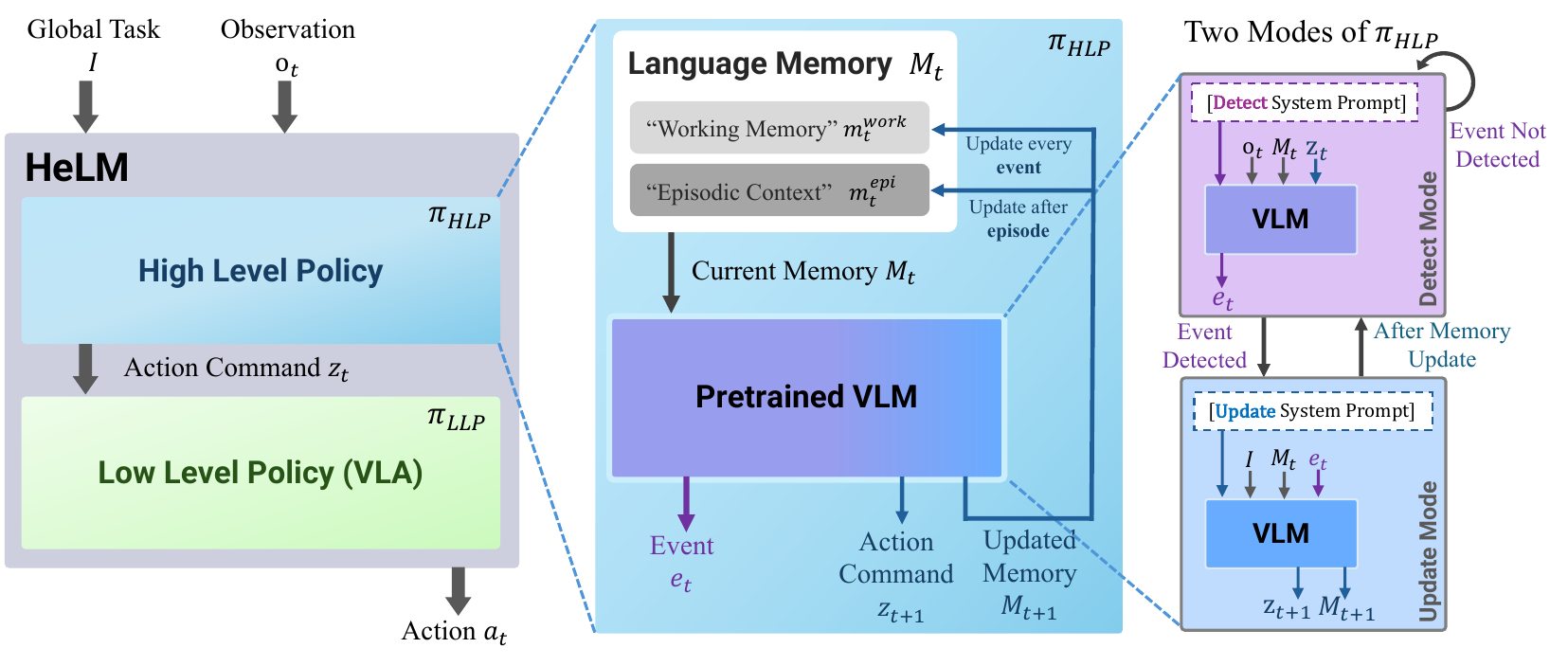}
    \caption{\textbf{Overall Architecture of HeLM.}
    The High Level Policy (HLP) integrates language memory with a pretrained VLM, which switches between a `detect mode' that monitors events and an `update mode' that updates these events into language memory.
    The Low Level Policy (LLP), a pretrained VLA, executes action commands generated by the HLP.}
    \label{fig:overall_architecture}
\end{figure*}

%% file: figure_tex/figure_3.tex
\begin{figure*}[t] %
    \centering
    \includegraphics[width=0.99\textwidth]{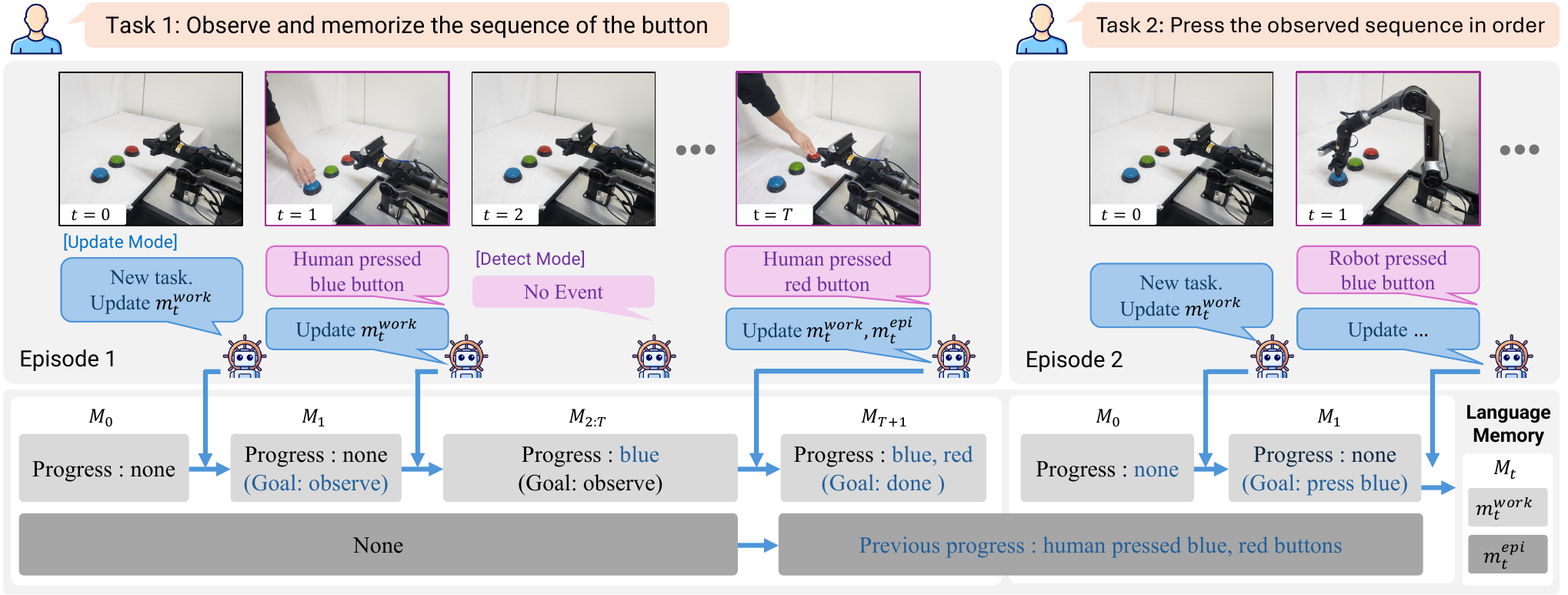}
    \caption{\textbf{Update / Detect mode of HLP in evaluation loop.} An example of transitions between Detect and Update modes and corresponding memory updates across intra- and inter-episode scenarios. Each action $a_t$ is generated based on $M_{t-1}$ and the current instruction.}
    \label{fig:evaluation}
\end{figure*}

%% file: section/experiments.tex
\input{figure_tex/figure_4}

\input{graph_tex/Intra_Episode}

\section{Experiments}
\label{sec:experiments}
Our experiments assess
(1) real-world performance on intra- and inter-episode tasks
(Sec.~\ref{sec:real_world}),
(2) the failure modes of HeLM and the baselines
(Sec.~\ref{sec:failure_analysis}),
(3) the contributions of language memory, HLP fine-tuning,
and two-stage inference through controlled ablations
(Sec.~\ref{sec:ablations}),
(4) generalization to held-out task configurations
(Sec.~\ref{sec:generalization}),
and (5) long-horizon robustness and computational efficiency
as task history grows
(Sec.~\ref{sec:scaling}).

\subsection{Experimental Setup}
\label{subsec:setup}

\noindent\textbf{Baselines.}
We compare HeLM against three baselines.
First, Keyframe-HLP is a variant of our HLP that stores visual keyframes of detected events instead of textual memory, isolating the effect of memory modality.
Second, PCMB-$\pi_0$ augments $\pi_0$ with the Perceptual-Cognitive Memory Bank (PCMB)~\cite{memoryvla}, which compresses and retrieves historical tokens via a memory module.
Finally, Naïve-$\pi_0$ ~\cite{pi0} is the base $\pi_0$ model trained end-to-end, without an explicit memory mechanism.
All baselines use the same $\pi_0$ action backbone fine-tuned on 150-200 demonstrations per task, controlling for low-level action capacity.
HeLM and Keyframe-HLP additionally share the same HLP backbone and event supervision, enabling a controlled comparison of memory modality.
Appendix~\ref{app:baselines} provides implementation details of the baselines, including the PCMB adaptation.

\noindent\textbf{Implementation of HeLM.}
We use Qwen2.5-VL-7B~\cite{bai2025qwen25vltechnicalreport} as the backbone of HLP, fine-tuned via QLoRA~\cite{dettmers2023qloraefficientfinetuningquantized} (rank $r{=}16$) on $\sim$130k frames generated via TaskSpec.
HLP runs every 10 LLP steps.
All real-world experiments use a 7-DoF AgileX PiPER arm with table/wrist-view RGB cameras.

\noindent\textbf{Evaluation Protocol.}
We report mean success rate per task over 15--45 trials (3--9 variants, 5 trials each).
\textit{Subtask success rate} measures the fraction of required subtasks completed, while order and termination errors are tracked separately to preserve failure-mode information.
\textit{Strict task success rate} requires all subtasks to be completed in the correct order with proper termination and is additionally reported for the intra-episode evaluation.
Appendix~\ref{app:success_metric} defines the subtask scoring rules and reports strict task success rates.

\input{table/Inter_Episode}

\subsection{Real-World Evaluation}
\label{sec:real_world}
\noindent\textbf{Intra-Episode.}
On intra-episode tasks (Fig.~\ref{fig:intra_episode_results}), HeLM substantially outperforms all baselines.
HeLM achieves 85.7\% average subtask success rate, compared to Keyframe-HLP (72.9\%), PCMB-$\pi_0$ (39.5\%) and Naïve-$\pi_0$ (16.7\%).
Under strict task success, HeLM retains 81.3\%, while Keyframe-HLP drops to 29.3\% and both PCMB-$\pi_0$ and Na\"ive-$\pi_0$ to 0\%.
The small gap between subtask and strict success for HeLM indicates reliable sequencing and termination, whereas the larger drops in the baselines reflect frequent order and termination failures (Sec.~\ref{sec:failure_analysis}).

\noindent\textbf{Inter-Episode.}
On inter-episode tasks (Tab.~\ref{tab:real_world_results_ie}), HeLM achieves 89.2\% average subtask success rate.
PCMB-$\pi_0$ and Naïve-$\pi_0$ are excluded as they reset internal states at episode boundaries, lacking support for cross-episode context; developing dedicated inter-episode baselines remains an open direction.
Keyframe-HLP degrades to 16.6\% average subtask success rate as visual keyframes accumulate across instruction boundaries.
In contrast, HeLM's episode context summarizes inter-episode information, enabling reliable progress tracking across instruction boundaries.

\subsection{Failure Analysis}
\label{sec:failure_analysis}
\noindent\textbf{Failure Categories.}
We categorize observed failures into four types:
Low-Level Execution (LLP fails physical execution),
Detection (HLP's visual event detection fails),
Order (subtask order violations),
and Termination (fail to stop after task completion or premature stopping).

\noindent\textbf{HeLM Failures.}
HeLM's failures are dominated by Detection (64.1\%) and Low-Level (31.2\%), while Order failures are rare (4.7\%) and Termination is not observed.
Detection failures are primarily false positives, especially for drawer-opening (A4, E4) and completion events (A3, A5); despite conservative prompting to suppress uncertain detections, these errors lead to incorrect memory updates that the framework cannot self-correct.
Low-Level failures arise in contact-sensitive manipulation (e.g., A1, A5, E1), where the LLP reaches the target region but fails to physically press or complete the interaction.

\noindent\textbf{Baseline Failures.}
Keyframe-HLP shows substantial Termination (33.3\%) and Detection (20.2\%) failures, reflecting confusion when visual keyframes accumulate.
PCMB-$\pi_0$ and Naïve-$\pi_0$ are dominated by Order and Termination failures across all tasks.
Overall, HeLM's failures are dominated by perception and low-level execution errors, highlighting these components as key targets for improvement.
A detailed breakdown of failure rates across individual tasks, along with representative failure cases, is provided in Appendix~\ref{app:failure_overview}.

\input{table/Libero-Mem-main}
\input{table/two_stage}
\subsection{Memory and Inference Ablations}
\label{sec:ablations}

We isolate key components of HeLM through complementary simulation and real-world ablations.

\noindent\textbf{Memory and HLP Finetuning on LIBERO-Mem.}
We evaluate HeLM on LIBERO-Mem~\cite{liberomem}, using $\pi_{0.5}$ as the LLP and Qwen2.5-VL-7B as the HLP which is shared across all ten tasks.
T1--T2 require short-horizon state tracking, T3--T8 require longer-term count, sequence, or relation tracking, and T9--T10 evaluate object occlusion.
We train a single HLP shared across all ten tasks using 100 demonstrations per task and evaluate 20 held-out rollouts per task.
We compare HeLM with $\pi_{0.5}$, a Memoryless HLP with the same hierarchy but no language memory, and an HLP without finetuning using the same memory interface.

As shown in Tab.~\ref{tab:libero_mem_main}, HeLM achieves 58.5\% average success rate compared with 36.5\% for $\pi_{0.5}$.
On the memory-intensive T3--T8 tasks, HeLM achieves 45.8\%, while $\pi_{0.5}$ achieves 10.0\% and both ablations achieve 0\%.
The Memoryless HLP repeatedly loses completed steps or counts, while the HLP without finetuning often fails to detect task completion or construct the appropriate memory state.
These results show that HeLM's gains cannot be explained by added hierarchical capacity alone and that HLP finetuning is important for task-specific grounding.
See more details in Appendix~\ref{sec:appendix_libero_mem}.

\noindent\textbf{Two-Stage Inference on Real-World Tasks.}
We compare HeLM with a one-stage variant that jointly predicts event completion, action command, and updated memory at every HLP call.
The one-stage variant achieves 67.7\% average subtask success, compared with 85.7\% for HeLM (Tab.~\ref{tab:two_stage_ablation}).
Its failures are mainly premature completion predictions and incorrect memory updates, with the largest degradation on A3.
These results support the benefit of the two-stage inference design.

\input{table/Zero_Shot}
\subsection{Compositional Generalization}
\label{sec:generalization}

\noindent\textbf{Compositional Generalization Experiment Setup.} 
We evaluate compositional generalization on novel task compositions of seen subtasks.
While the LLP is fine-tuned to execute specific action commands, the HLP receives no additional finetuning on the held-out variants.
The evaluation varies unseen counts, longer action orders, unseen objects, and unseen compositions of previously learned subtasks.
Appendix~\ref{sec:appendix_zeroshot_tasks} describes the design of the held-out variants used in this evaluation.

\noindent\textbf{Quantitative Analysis.}
As shown in Tab.~\ref{tab:compositional_generalization}, HeLM achieves 78\% average subtask success rate.
Keyframe-HLP achieves 65.7\% but fails to detect termination in A1', A4', and A5'.
These results support HeLM's ability to compose learned subtasks beyond the training configurations.

\input{graph_tex/Latency_VRam}
\subsection{Memory Scaling and Overhead: Text vs. Visual Memory}
\label{sec:scaling}
We compare textual memory against a visual keyframe baseline (Keyframe-HLP) on long-horizon performance and computational efficiency.
For this comparison, both HeLM and Keyframe-HLP use the same Qwen2.5-VL-7B HLP backbone.

\noindent\textbf{Long-Horizon Robustness.}
As shown in Fig.~\ref{fig:task_length_results}, HeLM maintains 100\% success on \textit{A1 long} as the repetition count increases to seven and remains at least 77\% on \textit{A2 long} as the required action sequence grows to seven subtasks.
Performance remains robust as the required task history grows within the evaluated range.

\noindent\textbf{Computational Efficiency.} 
HeLM is initially slower due to autoregressive text generation, but becomes 2.5$\times$ faster at long horizons as visual tokens accumulate in Keyframe-HLP (see Fig.~\ref{fig:task_length_results}).
Keyframe-HLP's peak VRAM usage reaches 28 GB for an 8-event episode, while HeLM remains under 6 GB.
This scaling advantage comes with additional HLP inference over a plain VLA.
Appendix~\ref{app:small_models} further evaluates smaller HLP and LLP backbones and their trade-off between subtask success and inference cost.

%% file: figure_tex/figure_4.tex
\begin{figure*}[t]
    \centering
    \includegraphics[width=\textwidth]{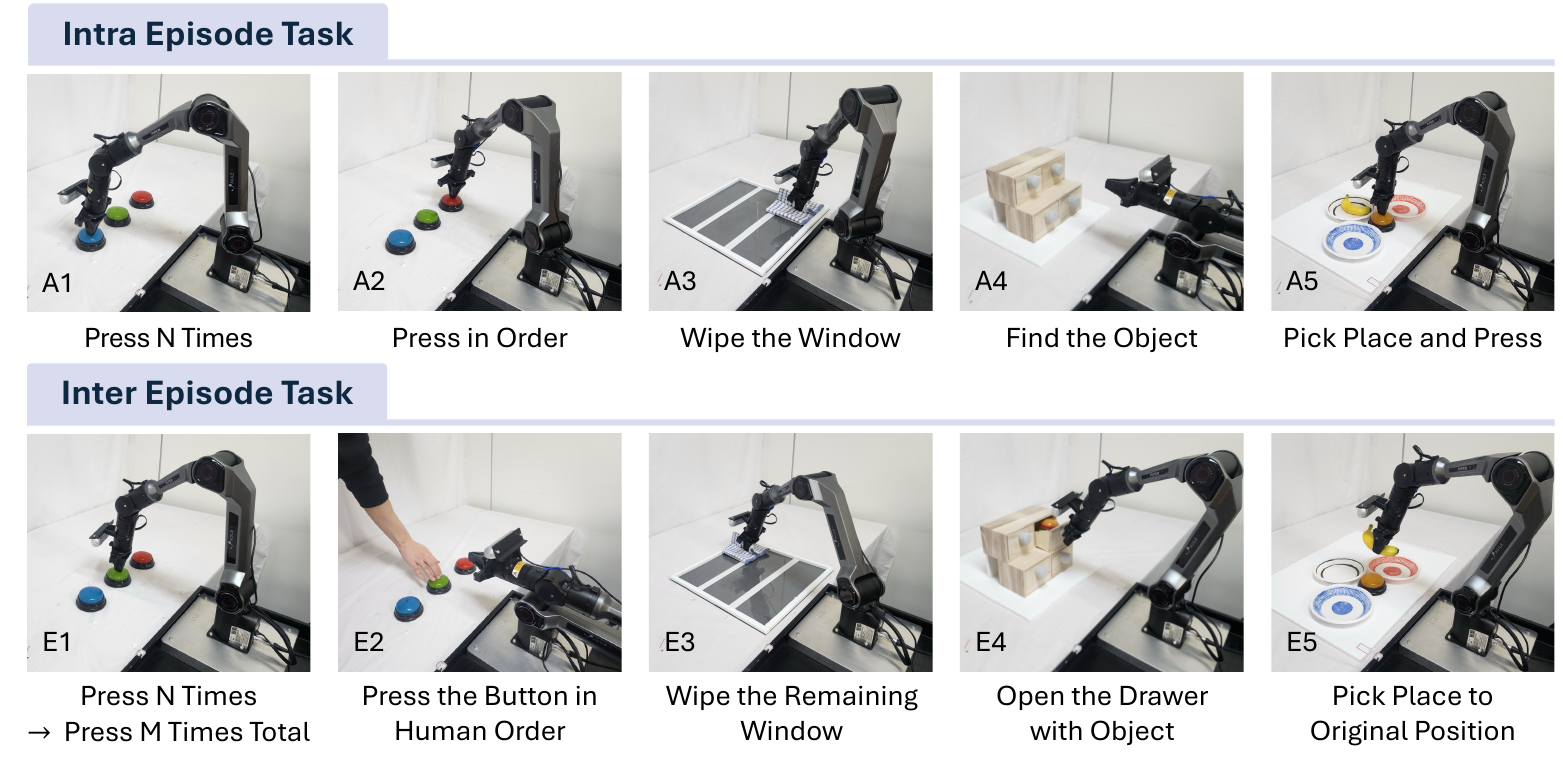}
    \caption{\textbf{The Real-World Evaluation Suite of HeLM.} Our tasks are categorized into two types based on their temporal memory requirements. (Top) Intra-episode tasks (A1–A5) evaluate working memory within a single instruction. (Bottom) Inter-episode tasks (E1–E5) assess episodic context across task boundaries.}
    \label{fig:experiment_task}
\end{figure*}

%% file: graph_tex/Intra_Episode.tex
\begin{figure*}[t]
    \centering

    \includegraphics[width=1\textwidth]{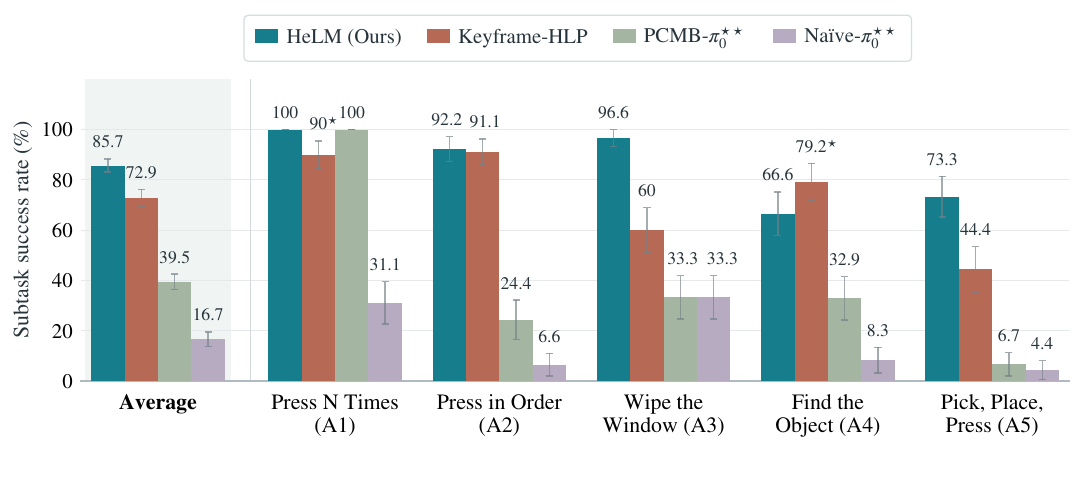}
    \vspace{-1cm}
    \caption{\textbf{Intra-episode real-world evaluation results.}
    HeLM achieves the highest average subtask success rate on five intra-episode tasks.
    In contrast, Keyframe-HLP fails to detect termination in A1, A4 (denoted by $\star$), while the baselines PCMB-$\pi_0$ and Naïve-$\pi_0$ (denoted by $\star\star$) fail both sequencing and termination across all trials. Under strict task success, HeLM retains 81.3\%, while Keyframe-HLP drops to 29.3\% and both PCMB-$\pi_0$ and Na\"ive-$\pi_0$ drop to 0\%.
    Bars report mean subtask success rate over 15–30 trials per task (3–6 variants, 5 trials each).
    Error bars denote one standard deviation across variants.}
    \label{fig:intra_episode_results}
\end{figure*}

%% file: table/Inter_Episode.tex
\begin{table}[t]
    \centering
    \caption{\textbf{Inter-episode real-world evaluation.}
    We report subtask success rates (\%) on tasks E1--E5,
    which require retaining task-relevant information across
    episode boundaries.
    Avg.\ denotes the mean across the five tasks.
    $\star$ indicates failure to terminate.
    }
    \label{tab:real_world_results_ie}
    \vspace{4pt}

    \small
    \setlength{\tabcolsep}{7pt}
    \renewcommand{\arraystretch}{1.15}

    \begin{tabular}{lcccccc}
        \toprule
        Method & E1 & E2 & E3 & E4 & E5 & Avg. \\
        \midrule
        Keyframe-HLP
        & $27.6^{\star}$ & 0.0 & 33.3
        & 0.0 & 22.2 & 16.6 \\
        \textbf{HeLM (Ours)}
        & \textbf{90.5} & \textbf{87.7} & \textbf{90.0}
        & \textbf{80.0} & \textbf{97.7} & \textbf{89.2} \\
        \bottomrule
    \end{tabular}
\end{table}

%% file: table/Libero-Mem-main.tex
\begin{table}[t]
    \centering
    \caption{\textbf{Memory and HLP fine-tuning ablations on LIBERO-Mem.}
    We report mean task success rates (\%) for three task groups
    and the average across all ten tasks.
    All configurations use $\pi_{0.5}$ ~\cite{intelligence2025pi05visionlanguageactionmodelopenworld} as the low-level policy.
    The memory ablation removes explicit language memory,
    while the fine-tuning ablation retains memory but uses
    an HLP without task-specific fine-tuning.
    T3--T8 require persistent temporal state tracking.
    }
    
    \vspace{4pt}
    \label{tab:libero_mem_main}

    \small
    \renewcommand{\arraystretch}{1.15}
    \setlength{\tabcolsep}{7pt}
    
    \begin{tabular}{lcccc}
        \toprule
        Method & T1--T2 & T3--T8 & T9--T10 & Avg. \\
        \midrule
        $\pi_{0.5}$
        & 65.0 & 10.0 & \textbf{87.5} & 36.5 \\
        \addlinespace[2pt]
        $\pi_{0.5}$ + HLP (w/o memory)
        & 27.5 & 0.0 & 32.5 & 12.0 \\
        $\pi_{0.5}$ + HLP + memory (w/o FT)
        & 5.0 & 0.0 & 37.5 & 8.5 \\
        \midrule
        \textbf{HeLM (Ours)}
        & \textbf{67.5} & \textbf{45.8}
        & \textbf{87.5} & \textbf{58.5} \\
        \bottomrule
    \end{tabular}
\end{table}

%% file: table/two_stage.tex
\begin{table}[t]
    \centering
    \caption{\textbf{Two-stage inference ablation on real-world tasks.}
    We report subtask success rates (\%) on intra-episode tasks A1--A5.
    The one-stage variant jointly predicts events, memory updates,
    and action commands, whereas HeLM separates event detection
    from event-triggered memory updates and command generation.
    }
    \label{tab:two_stage_ablation}
    \vspace{4pt}

    \small
    \setlength{\tabcolsep}{6pt}
    \renewcommand{\arraystretch}{1.15}

    \begin{tabular}{lcccccc}
        \toprule
        Method & A1 & A2 & A3 & A4 & A5 & Avg. \\
        \midrule
        HeLM w/ one-stage inference
        & \textbf{100.0} & 81.5 & 22.2
        & \textbf{86.8} & 48.1 & 67.7 \\
        \textbf{HeLM (Ours)}
        & \textbf{100.0} & \textbf{92.2} & \textbf{96.6}
        & 66.6 & \textbf{73.3} & \textbf{85.7} \\
        \bottomrule
    \end{tabular}
\end{table}

%% file: table/Zero_Shot.tex
\begin{table}[t]
    \centering
    \caption{\textbf{Compositional generalization on real-world tasks.}
    We report subtask success rates (\%) on held-out variants
    A1$^{\prime}$--A5$^{\prime}$ of the intra-episode tasks.
    These variants evaluate unseen configurations of learned
    subtasks without additional HLP fine-tuning.
    $\star$ indicates failure to terminate.
    }
    \label{tab:compositional_generalization}
    \vspace{4pt}

    \small
    \setlength{\tabcolsep}{7pt}
    \renewcommand{\arraystretch}{1.15}

    \begin{tabular}{lcccccc}
        \toprule
        Method
        & A1$^{\prime}$ & A2$^{\prime}$ & A3$^{\prime}$
        & A4$^{\prime}$ & A5$^{\prime}$ & Avg. \\
        \midrule
        Keyframe-HLP
        & $\mathbf{100.0}^{\star}$ & 63.9 & \textbf{77.8}
        & $38.9^{\star}$ & $48.1^{\star}$ & 65.7 \\
        \textbf{HeLM (Ours)}
        & \textbf{100.0} & \textbf{95.0} & 67.0
        & \textbf{67.0} & \textbf{61.0} & \textbf{78.0} \\
        \bottomrule
    \end{tabular}
\end{table}

%% file: graph_tex/Latency_VRam.tex
\begin{figure}[t]
    \centering
    \includegraphics[width=\linewidth]{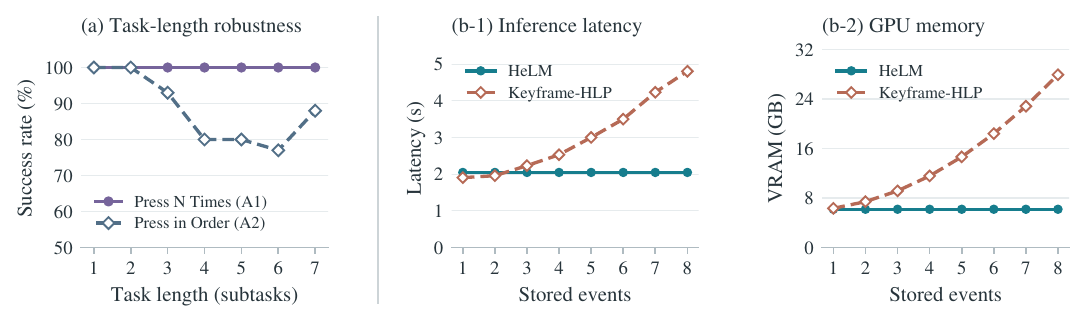}
    \caption{\textbf{Task-length robustness and memory scaling.}
    (a) HeLM success rates on extended variants of
    A1 (Press N Times) and A2 (Press in Order)
    as the number of required subtasks increases.
    (b-1, b-2) Inference latency and GPU memory usage
    versus the number of stored events for HeLM and Keyframe-HLP.
    }
    \label{fig:task_length_results}
\end{figure}

%% file: section/conclusion.tex
\section{Limitations, Future Work, and Conclusion} 
\label{sec:conclusion}
\noindent\textbf{Limitations and Future Work.}
Several limitations remain.
First, HeLM relies on predefined task-relevant events through TaskSpec, which requires task-specific event definitions and does not address open-ended discovery of what should be remembered.
Second, our evaluation focuses on discrete events and does not cover continuous state tracking such as liquid levels or precise geometric history.
Third, the current synchronous chunk-level HLP invocation introduces inference overhead and may miss transient events that occur entirely within an LLP action chunk.
Fourth, the benefit of separating working and episodic memory, relative to using a single persistent language memory, remains to be evaluated.
These limitations motivate more adaptive event discovery, alternative memory organizations, and higher-frequency or asynchronous high-level inference.

\vspace{0.1cm}
\noindent\textbf{Conclusion.}
We introduced HeLM (Hierarchical Explicit Language Memory) for non-Markovian manipulation, where current observations alone are insufficient for action selection.
HeLM maintains task-relevant history in explicit language and decouples high-level memory management from low-level action generation.
Across intra- and inter-episode tasks, HeLM improves long-horizon progress tracking and enables information from earlier interactions to guide later decisions.
These results highlight memory representation as an important design dimension for robot policies and show that language provides a compact and interpretable interface for maintaining task-relevant history over time.

%% file: section/appendix.tex
\section{HeLM Implementation}
\label{app:helm_implementation}
This section provides implementation details of HeLM including: dataset generation method using taskspec, system prompt for two modes of HeLM HLP, and train hyperparameters.

\input{figure_tex/figure_appendix_data_gen}
\vspace{0.3cm}
\noindent\textbf{Train Dataset Generation.}
To train the HLP model with robust event-driven reasoning, we developed a TaskSpec-driven data generation pipeline.
This approach leverages a key architectural advantage of HeLM: unlike latent memory models~\cite{memoryvla, liberomem} that require full video histories for training, HeLM's explicit text memory decouples the current decision from the raw visual past.
This allows us to treat each training sample as an independent tuple of observation and memory state $\langle o_t, M_t \rangle$.

We define \textit{TaskSpec} as a schema that maps specific stages of non-Markovian tasks (e.g., ``count: 2'') to corresponding event frames and valid memory states.
As in Fig~\ref{fig:appendix_taskspec_example}, a TaskSpec explicitly specifies the valid action commands, detected events, and corresponding memory updates at each stage, ensuring consistent alignment between visual observations and language memory states.
Using this schema, we synthetically generate a dataset without manually collecting data for every task variant.

By sampling image frames from raw episodes and pairing them with language memories defined in TaskSpec, we create massive variations of logic-intensive scenarios.
This effectively teaches the model to handle language memory; intra-episode sequential logic (updating $m^{\text{work}}$) and inter-episode context transitions (promoting $m^{\text{work}}$ to $m^{\text{epi}}$) by exposing it to diverse language memory-visual observation combinations.
Crucially, events act as discrete anchors that trigger memory updates, allowing the model to learn when and how memory should change based on visual evidence rather than continuous observation streams.

\noindent\textbf{TaskSpec Generation.}
TaskSpec authoring involves defining the logical schema (event vocabulary, action commands, memory transition rules) per task.
Task variants are generated by GPT using the hand-written TaskSpec as a template, with manual verification for schema consistency (Fig.~\ref{fig:appendix_taskspec_generation_prompt}).
The main manual cost is event frame annotation, approximately 5 minutes per subtask (covering ~30 episodes).

{\input{table/hyperparameter}}
\vspace{0.3cm}
\noindent\textbf{System Prompt.}
HeLM ensures memory stability through a two-stage update mechanism that triggers only upon detecting a semantic event.
The two modes of the HLP share the same VLM parameters but are functionally distinguished through task specific system prompts.
Fig~\ref{fig:appendix_detect_prompt} and Fig~\ref{fig:appendix_update_prompt} show the full system prompts used to train the HLP in \texttt{DETECT} and \texttt{UPDATE} modes.

\vspace{0.3cm}
\noindent\textbf{Train Hyperparameters.}
We train HeLM using QLoRA, where the base model weights are quantized to 4-bit NF4 with double quantization, while all computations, LoRA adapters, and activations are performed in BF16.
The following Tab~\ref{tab:train_hparams} shows details of training hyperparameters used for HeLM.

\input{section/appendix/Task_Detail}

\input{section/appendix/Failure_Analysis}
\input{section/appendix/Experiment_Detail}

\section{Additional Experiments}
\label{app:additional_experiment}

This section presents additional experimental analyses examining (i) HeLM's scalability to longer task sequences, and (ii) the framework's compatibility with smaller, more efficient backbone models. These analyses highlight trade-offs between reliability, latency, and resource efficiency.

\input{table/long_task}

\subsection{Long-Horizon Robustness.}
We evaluate HeLM's scalability to longer task sequences using extended variants of A1 (Press N Times) and A2 (Press in Order). As shown in Table~\ref{tab:long_task_results_ia}, HeLM maintains success rates above 80\% even at task length 7 on both tasks (A1: 100\%, A2: 88\%), demonstrating that explicit language memory remains stable as task complexity increases.

This robustness stems from HeLM's compression of historical context into discrete textual tokens rather than accumulating high-dimensional visual features. By filtering out background variation and capturing only semantically meaningful events, the memory representation avoids the overhead and noise associated with raw observation history in long-horizon settings.

\input{table/Ablation_model}

\subsection{Model Generalizability and Scalability.}
\label{app:small_models}
We evaluate HeLM's compatibility with smaller backbone models by replacing the high-capacity models with significantly more efficient alternatives. The base HeLM uses Qwen2.5-VL (7B) as the High-Level Policy (HLP) and $\pi_0$ (3.5B) as the Low-Level Policy (LLP). In the efficient variant, these are replaced with SmolVLM (500M)~\cite{marafioti2025smolvlmredefiningsmallefficient} and SmolVLA (450M)~\cite{shukor2025smolvlavisionlanguageactionmodelaffordable}, respectively.

As shown in Table~\ref{tab:ablation_model}, this transition results in a reduction in the average success rate from 85.7\% to 58.7\%, but yields a substantial improvement in computational efficiency. Specifically, the peak VRAM consumption is reduced from 13.18 GB to 2.3 GB, and the average end-to-end latency is shortened from 1.9s to 0.7s. These results indicate that HeLM can be deployed on resource-constrained edge devices, at the cost of a predictable performance trade-off.

Detailed error analysis reveals that the performance degradation is primarily concentrated in the logical updates of the HLP. The reduced parameter scale of the 500M model leads to frequent failures during the \texttt{UPDATE} mode, including generating invalid YAML formats or updating multiple state variables simultaneously, which causes the agent to skip critical subtasks. This tendency is particularly evident in complex scenarios such as A4 and A5, where the increased length of global instructions and subtasks (see Tab.~\ref{tab:task_details_appendix}) hinders the construction of accurate language memory. Nevertheless, the framework's core memory mechanism—maintaining and updating language memory across subtasks—remains functional even at this reduced scale, suggesting that optimizing the memory representation into more concise and structured formats could recover success rates while preserving resource efficiency.

\input{table/Libero-mem}
\subsection{Simulation Evaluation on LIBERO-Mem}
\label{sec:appendix_libero_mem}

As a complement to our main real-robot experiments, we ran HeLM on the LIBERO-Mem benchmark~\cite{liberomem} to compare with published memory-augmented baselines under simulation.

\vspace{0.3cm}
\noindent\textbf{Benchmark Overview.}
LIBERO-Mem consists of 10 simulation tasks designed to evaluate four dimensions of object-centric memory: Object Motion (OM), Object Sequence (OS), Object Relations (OR), and Object Occlusion (OO).
T1 and T2 (OM) test recall of the last action---since visually identical states correspond to different required actions (e.g., lifting vs placing), the agent must remember its most recent action.
T3--T6 (OM + OS) extend this with repeated pick-and-place actions (3, 3, 5, and 7 times respectively); the agent must track how many times each object has been manipulated since visual cues alone are insufficient.
T7 and T8 (OM + OR) test relational tracking through bowl-swapping tasks using an empty buffer plate. Bowls are placed at randomized initial positions, so each variation presents different geometric configurations; the agent must maintain the temporal order of object relations across multi-step rearrangements.
T9 and T10 (OM + OO) test object occlusion by requiring the agent to place a target object into one of two visually identical baskets and then manipulate the correct basket; the intended occlusion forces the agent to remember which basket contains the target.

\vspace{0.3cm}
\noindent\textbf{Benchmark Limitations.} 
We observed two issues during evaluation that may affect interpretation of the results:
(i) For T9 and T10, the wrist-mounted camera observes the basket's interior even after the target object is placed inside, so the intended occlusion does not effectively take place. This allows memoryless policies to verify object location visually rather than relying on memory, which likely explains why $\pi_{0.5}$ alone achieves high performance on these tasks (85--90\%).
(ii) Certain initial state configurations exhibit object spawn errors---for instance, demos 108 and 115 spawn a bottle pre-placed on its target plate, conflicting with the BDDL goal state and causing systematic failure across multiple tasks regardless of method. These environment-side issues affect all evaluated methods (memoryless baselines and HeLM) equally and are independent of method design.
We therefore interpret these results as complementary simulation evidence, taking the observed benchmark limitations into account.

\vspace{0.3cm}
\noindent\textbf{Setup.} 
We use $\pi_{0.5}$ as the LLP and Qwen2.5-VL-7B as the HLP. Both the vanilla $\pi_{0.5}$ baseline and the HeLM LLP are trained on 100 demonstrations per task under a multitask training regime, following the train/validation split provided by the LIBERO-Mem benchmark (100 training + 20 validation trajectories per task). The LLP is fully fine-tuned starting from a LIBERO-pretrained $\pi_{0.5}$ checkpoint (40k--55k training steps, effective batch size 24, cosine LR schedule with peak $2.5 \times 10^{-5}$), and the HLP is fine-tuned with QLoRA (rank 16, effective batch size 32). HeLM's LLP training data is generated by a rule-based pipeline that performs subtask splitting, event annotation, and TaskSpec generation on the demonstrations, following the same procedure used for our real-robot experiments. All training is performed on 2$\times$NVIDIA A100 GPUs. Evaluation is performed on the held-out validation set: each task is evaluated over 20 rollouts using the initial states of the validation trajectories, consistent with the original benchmark protocol.

We chose $\pi_{0.5}$ over $\pi_0$ as the LLP because $\pi_0$ failed to learn even individual subtasks under our training regime---likely due to weaker language grounding for fine-grained subtask instructions. Using $\pi_0$ as the LLP would have caused HeLM's evaluation to be dominated by LLP execution failures rather than memory-related failures, preventing meaningful assessment of the memory mechanism. For the same reason, we additionally report $\pi_{0.5}$ alone as a memoryless baseline, since the original benchmark only reports results with $\pi_0$.

\vspace{0.3cm}
\noindent\textbf{Results.} 
As shown in Tab.~\ref{tab:libero_mem}, HeLM achieves 58.5\% average success rate across the 10 LIBERO-Mem tasks, substantially outperforming the memoryless $\pi_{0.5}$ baseline (36.5\%) and published memory-augmented baselines (Naive E-SlotSSM: 14.8\%). The improvement is most pronounced on memory-intensive tasks (T3--T8), where memoryless baselines fail to reason over temporal dependencies.

\vspace{0.3cm}
\noindent\textbf{Discussion.} 
The per-task pattern reveals how language memory contributes across different memory dimensions.

\textit{Object Motion (T1--T2).} These short-horizon tasks involve minimal memory requirements (recall of the last action). HeLM outperforms $\pi_{0.5}$ on T1 (75\% vs 50\%), while $\pi_{0.5}$ performs better on T2 (80\% vs 60\%), reflecting that the gap between methods varies on tasks where memory plays a limited role.

\textit{Object Sequence (T3--T6).} HeLM preserves past actions in language memory and uses them to drive correct sequential execution, yielding substantial gains over the memoryless $\pi_{0.5}$ (e.g., T3: 75\% vs 15\%, T5: 65\% vs 15\%). As the required repetition count grows (3 to 7 times), success rate decreases overall (T3: 75\%, T6: 25\%) due to accumulated LLP execution failures over longer horizons, even when the HLP's memory reasoning remains correct. This pattern is consistent with our main paper's observation that perception and execution---rather than memory reasoning---become the dominant bottleneck as horizons extend.

\textit{Object Relations (T7--T8).} These tasks place bowls at randomized positions and require rule-based swaps using an empty buffer plate.
Each variation produces different geometric configurations, making them particularly challenging.
Whether discrete language memory can preserve geometric relations is an open question for language-as-memory frameworks and a question this work investigates.
HeLM's success on these complex multi-step tasks (T7: 45\%, T8: 10\%) suggests that language memory, combined with the current visual observation, can partially preserve sufficient geometric information for relational reasoning---even without storing explicit positional encodings.

\textit{Object Occlusion (T9--T10).} As noted in the benchmark limitations, the wrist camera circumvents the intended occlusion in these tasks, effectively reducing them to memoryless settings. Both $\pi_{0.5}$ and HeLM achieve high performance (85--90\%), with the agreement between the two methods reflecting the LLP's underlying execution capability when memory is not the bottleneck.

Overall, the per-task results are consistent with our main paper's claim that explicit language memory provides stable performance on memory-sensitive non-Markovian tasks, while remaining transparent about tasks where the benchmark's intended challenges may not be fully realized.

\newpage
\input{figure_tex/figure_detect_mode_system_prompt}
\input{figure_tex/figure_update_mode_system_prompt}

\newpage
\input{figure_tex/figure_appendix_task}

\newpage
\input{figure_tex/figure_appendix_taskspec}

\clearpage
\input{figure_tex/figure_appendix_llm_generated_taskspec}

\newpage
{\input{figure_tex/figure_appendix_task_example}

%% file: figure_tex/figure_appendix_data_gen.tex
\begin{wrapfigure}{r}{0.6\columnwidth}
    \centering
    \includegraphics[width=0.58\columnwidth]{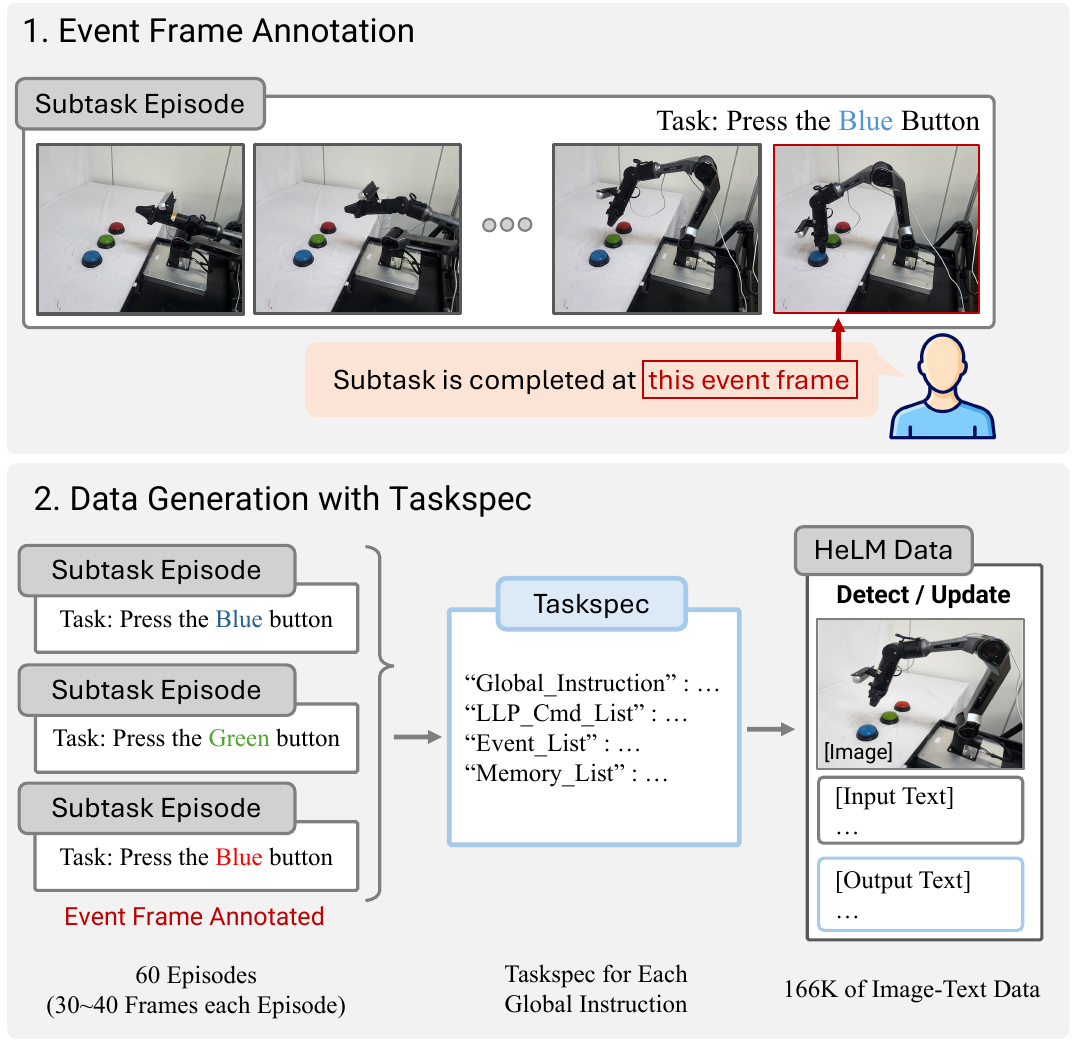} 
    \caption{\textbf{Generation of HeLM data.} Instead of recording multiple episodes of long-horizon non-Markovian tasks, we use a task configuration list to logically stitch together short subtask episodes.}
    \label{fig:appendix_data_gen}
\end{wrapfigure}

%% file: table/hyperparameter.tex
%
\begin{table}[t]
\centering
\captionof{table}{\textbf{Training hyperparameters used for HeLM.} We fine-tune Qwen2.5-VL (7B) as the HLP for our model.}
\label{tab:train_hparams}
\renewcommand{\arraystretch}{1.1}
\setlength{\tabcolsep}{6pt}
\begin{tabularx}{\linewidth}{@{}p{4cm} X@{}}
\toprule
\textbf{Hyperparameter} & \textbf{Value} \\
\midrule
Base model & Qwen2.5-VL-7B-Instruct \\
Optimizer & AdamW \\
Learning rate & $2\times 10^{-5}$ \\
Weight decay & 0.0 \\
Scheduler & Trainer default \\
LR warmup & warmup\_ratio = 0.03 \\
Quantization & QLoRA: 4-bit NF4 \\
\midrule
Batch size &
\makecell[l]{
total=8 \\
detect\_pos : 2 \\
detect\_neg : 2 \\
update\_intra : 2 \\
update\_transition : 2
} \\
Training steps & 3000 steps \\
Gradient accumulation & 1 \\
\midrule
LoRA rank $r$ & 16 \\
LoRA $\alpha$ & 32 \\
LoRA dropout & 0.05 \\
LoRA target modules & \texttt{q\_proj, k\_proj, v\_proj, o\_proj, gate\_proj, up\_proj, down\_proj} \\
Attention implementation & SDPA \\
\midrule
Image resolution (raw) & 480$\times$640 \\
\# images per sample & 1 \\
\midrule
Train data composition &
\makecell[l]{
Total = 166{,}590 (Train: ~130k / Validation: ~36k)\\
detect\_neg: 104{,}425 (62.7\%)\\
detect\_pos: 30{,}416 (18.3\%)\\
update\_intra: 30{,}416 (18.3\%)\\
update\_transition: 1{,}333 (0.8\%)
} \\
\midrule
Hardware & Single NVIDIA RTX A6000 GPU \\
Training time & $\sim$15 hours for 3000 steps \\
\bottomrule
\end{tabularx}
\end{table}

%% file: section/appendix/Task_Detail.tex
\input{table/appendix_task_list}

\section{\textbf{Task Details}}

\subsection{\textbf{Details of Intra- and Inter-episode Task}}
\label{app:task_detail}
This section details the intra-episode tasks (A1--A5) and inter-episode tasks (E1--E5), including the exact episode variants used for each task, as illustrated in Tab~\ref{tab:task_details_appendix} and Fig~\ref{fig:appendix_tasks}.

\vspace{0.3cm}
\noindent{A1: Press $N$ Times.}

\noindent The scene contains a single {blue button}. The agent is instructed to press the blue button exactly $N$ times and then stop. We evaluate three scenarios with $N \in \{1,2,3\}$.

\vspace{0.3cm}
\noindent{A2: Press in Order.}

\noindent The scene contains three colored buttons, {blue (B)}, {green (G)}, and {red (R)}. The agent is instructed to press the buttons in a specified order and then stop. We evaluate six scenarios corresponding to all variations:
\texttt{BGR}, \texttt{BRG}, \texttt{GBR}, \texttt{GRB}, \texttt{RGB}, \texttt{RBG}.

\vspace{0.3cm}
\noindent{A3: Wipe the Window.}

\noindent The scene contains a window divided into three regions: {bottom}, {middle}, and {top}. The agent is instructed to wipe the window in all variations: {bottom, middle, top} and then stop.
\vspace{0.3cm}

\noindent{A4: Find the Object.}

\noindent The scene contains four drawers located at {top-left}, {top-right}, {bottom-right}, and {bottom-left}. A toy {hamburger} is hidden in exactly one drawer. The agent is instructed to find the toy hamburger by opening drawers in a fixed search order: starting from {top-left} and continuing {clockwise} (top-left $\rightarrow$ top-right $\rightarrow$ bottom-right $\rightarrow$ bottom-left) until the hamburger is found. We evaluate four scenarios, each differentiated by the hamburger's hidden location (one per drawer).

\vspace{0.3cm}
\noindent{A5: Pick, Place, and Press.}

\noindent The scene contains three dishes---{blue}, {red}, and {white}---and a {center orange button}. A {banana} must be moved between dishes according to the instruction, with a button press occurring between two transfers. We evaluate six scenarios corresponding to all permutations of the 
transfer path across the three plates (Red, Blue, White): 
BRW, BWR, RBW, RWB, WBR, and WRB.

\vspace{0.3cm}
\noindent{E1: Press $N$ Times $\rightarrow$ Press $M$ Times Total.}

\noindent In the first episode, the robot presses the blue button $N$ times same as task A1.
Then, the agent is given a new instruction to press the button a total of $M$.
To succeed, the agent must recall the specific count $N$ to infer that exactly $M-N$ additional presses are required. 
We evaluate across nine scenarios with $N \in \{1,2,3\}$ and $M-N \in \{1,2,3\}$, ranging from simple increments like \texttt{1+1 (Total 2)} to more complex sequences such as \texttt{3+3 (Total 6)}.

\vspace{0.3cm}
\noindent{E2: Press the Button in Human Order.}

\noindent In the first episode, the human demonstrates a button-press sequence using the three colored buttons {B}, {G}, and {R}, and the agent is required to reproduce the demonstrated order. We evaluate six scenarios corresponding to:
\texttt{BGR}, \texttt{BRG}, \texttt{GBR}, \texttt{GRB}, \texttt{RGB}, \texttt{RBG}.

\vspace{0.3cm}
\noindent{E3: Wipe the Remaining Window.}

\noindent This task evaluates the ability to complete the remaining region(s) given partial progress.
The window is divided into {bottom (B)}, {middle (M)}, and {top (T)}.
In each task, one or two regions are wiped in the first episode, and the agent is instructed to wipe the remaining region in the next episode.
We evaluate six scenarios covering all variants (e.g., \texttt{BM + T}).

\vspace{0.3cm}
\noindent{E4: Open the Drawer with Object.}

\noindent In the first episode, the human opens the drawer with the toy hamburger, and the human then closes the drawer after the agent detects the toy. The agent is then instructed to directly open the same drawer (as in A4) that contains the hamburger (without sequential search). 
We evaluate four scenarios, one for each drawer location.

\vspace{0.3cm}
\noindent{E5: Pick, Place to Original Positions.}

\noindent This task extends task A5 by requiring the banana to be returned to its original positions after the initial two transfers. 
In the first episode, the agent executes one of the A5 procedures ({BRW}, {RBW}, {WBR}), and then performs an additional return sequence to restore original placements, using:
{WB}, {WR}, {RW}, {RB}, {BR}, and {BW}, respectively.
We evaluate six scenarios corresponding to the three initial A5 variants.

\subsection{\textbf{Details of Compositional Generalization Tasks}}
\label{sec:appendix_zeroshot_tasks}
Our compositional HLP generalization evaluation consists of variants (A1$^\prime$--A5$^\prime$) corresponding to five intra-episode tasks (A1--A5), each containing 3 variants.

\vspace{0.3cm}
\noindent\textbf{A1$^\prime$: Press $N$ Times.}
The agent is evaluated on values of $N$ beyond those seen during training (e.g., training includes up to three presses, while testing requires more, such as five presses), measuring whether it can extend counting-based sequential logic to longer horizons.

\vspace{0.3cm}
\noindent\textbf{A2$^\prime$: Press in Order.}
The agent is required to execute longer plans by repeating or concatenating specified sequences (e.g., RGBGBR), testing whether the HLP can extend ordered plans without task-specific fine-tuning.

\vspace{0.3cm}
\noindent\textbf{A3$^\prime$: Wipe the Window.}
The agent wipes window regions according to an instructed spatial order. While training uses a canonical order (Bottom $\rightarrow$ Middle $\rightarrow$ Top), A3$^\prime$ permutes the regions into novel orderings, evaluating generalization to unseen subtask sequences.

\vspace{0.3cm}
\noindent\textbf{A4$^\prime$: Find the Object.}
The agent must locate a target object specified by a semantic description among candidate containers. In A4$^\prime$, the target is replaced with an object not seen during training, testing whether the HLP can ground novel object concepts from language alone.

\vspace{0.3cm}
\noindent\textbf{A5$^\prime$: Pick, Place, and Press.}
The agent executes a multi-step procedure of picking an object, placing it on a receptacle, and pressing a button. A5$^\prime$ alters the procedure order (e.g., changing the order of object transfers), evaluating whether the agent can recombine familiar skills into unseen composite sequences.

%% file: table/appendix_task_list.tex
\begin{table*}[t]
\centering
\small
\renewcommand{\arraystretch}{1.4}
\caption{\textbf{Detailed description of the Real-World Evaluation Suite.}
The suite consists of five intra-episode tasks (A1--A5) for working memory and five inter-episode tasks (E1--E5) for episodic context, covering a diverse range of non-Markovian reasoning challenges. Each inter-episode task contains two language instructions.}
\label{tab:task_details_appendix}
\begin{tabularx}{\textwidth}{@{}c l >{\raggedright\arraybackslash}X c >{\raggedright\arraybackslash}p{2.8cm}@{}}
\toprule
\textbf{ID} & \textbf{Task Name} & \textbf{Language Instruction Template} & \textbf{Vars.} & \textbf{Variation Type} \\
\midrule
\rowcolor{gray!10}
\multicolumn{5}{l}{\textit{Intra-episode Tasks (Working Memory)}} \\
A1 & Press $N$ Times & Press the blue button $N$ times. & 3 & $N \in \{1,2,3\}$ \\
A2 & Press in Order & Press the [Order] buttons in order. & 6 & Order (B, G, R) \\
A3 & Wipe the Window & Wipe the [Order] part of the window in order. Each part of the window is wiped from left to right. & 6 & Order (bottom, middle, top) \\
A4 & Find the Object & Open the drawers in the following order to find the toy hamburger, and stop when you find it in the drawer: top\_left, top\_right, bottom\_right, bottom\_left. & 4 & Target Location \\
A5 & Pick, Place, Press & Pick and place the banana from the [A] plate to the [B] plate, press the orange button, and then pick and place the banana from the [B] plate to the [C] plate. & 6 & Transfer Path (Red, Blue, White) \\
\midrule
\addlinespace[0.5em]
\rowcolor{gray!10}
\multicolumn{5}{l}{\textit{Inter-episode Tasks (Episodic Context)}} \\
E1 & Cumulative Press & Press the blue button $N$ times. $\rightarrow$ Press the blue button $M$ times in total. & 9 & $N, M-N \in \{1,2,3\}$ \\
E2 & Human-Order Press & Observe the sequence of buttons pressed by the person. $\rightarrow$ Press the buttons in the same order as observed before. & 6 & Demonstrated Order (B, G, R) \\
E3 & Remaining Wipe & Wipe the [Order] parts of the window in order. Each part of the window is wiped from left to right. $\rightarrow$ Wipe the remaining part of the window. Each part of the window is wiped from left to right. & 6 & Order (bottom, middle, top) \\
E4 & Direct Retrieval & Observe in which drawer the toy hamburger is located. $\rightarrow$ Based on memory, find the drawer with the toy hamburger. & 4 & Target Location \\
E5 & Original Restore & Pick and place the banana from the [A] plate to the [B] plate, then from the [B] plate to the [C] plate. $\rightarrow$ Pick and place the banana from current plate to its original plate. & 6 & Transfer Path (Red, Blue, White) \\
\bottomrule
\end{tabularx}
\end{table*}

%% file: section/appendix/Failure_Analysis.tex
\input{figure_tex/figure_appendix_failure_cases}
\section{\textbf{Failure Analysis}}
\label{app:failure_overview}
We analyze failure cases of HeLM and baselines on both intra-episode and inter-episode tasks.
We categorize failures into four types as shown in Fig.~\ref{fig:appendix_failure_cases}:

\vspace{0.1cm}
\noindent(i) \textit{Low-Level Failure}: The agent produces actions aligned with the action command but fails to physically achieve the intended outcome.

\vspace{0.1cm}
\noindent(ii) \textit{Detection Failure}: Incorrect event detection, including false positives and false negatives. This category applies only to event-detection-based methods (HeLM and Keyframe-HLP).

\vspace{0.1cm}
\noindent(iii) \textit{Order Failure}: The agent fails to follow the required subtask order in order-sensitive tasks.

\vspace{0.1cm}
\noindent(iv) \textit{Termination Failure}: The agent fails to stop after task completion and continues executing actions.

\vspace{0.3cm}
\noindent\textbf{Failure Analysis. }
Tab.~\ref{tab:detailed_failure_analysis} shows that HeLM's failures across intra-episode tasks (A1--A5) are dominated by detection failures (9.1\% of trials) and low-level failures (4.4\%), while order violations are rare (0.7\%) and termination failures are not observed. Among HeLM's failure cases, detection and low-level errors together account for the majority of failure types.

Qualitatively, HeLM rarely fails in short-horizon intra-episode tasks such as A1 and A3, indicating that its reasoning and memory are generally stable within an episode.
When failures occur, they concentrate in two cases.
Low-level failures mainly arise in contact-sensitive manipulation, where the high-level reasoning is correct but the low-level policy does not produce the precise action.
For instance, in A1, A5, and E1, the LLP reaches the target region but often fails to physically press or complete the interaction, despite correct detection and memory states.
Detection failures are primarily false positives, especially for drawer-opening (A4) and completion (A3, A5) events.
In A4 and E4, the dominant failure case is detecting the drawer as opened even when it has not been opened.
This distinction is important for recovery: while repeated execution attempts can eventually resolve low-level failures when no progress event is detected, false-positive detections lead to incorrect next action commands.
Overall, these results suggest that HeLM's remaining bottlenecks are (i) contact-accurate low-level execution and (ii) robust event detection under occlusion.

As summarized in Tab.~\ref{tab:detailed_failure_analysis}, Keyframe-HLP exhibits a substantially higher overall failure rate, with termination failures (33.3\%) and detection failures (20.2\%) as the dominant categories, along with order failures (4.6\%) and low-level failures (2.2\%).
We also observe incorrect event detection (before task completion) in tasks such as A3, where completion is detected from an early keyframe.
These issues become more severe in inter-episode settings, where the method frequently fails to terminate or to maintain progress memory consistently across episodes.

For $\pi_0$-based baselines (PCMB-$\pi_0$ and Na\"ive-$\pi_0$), failures are dominated by termination and order failures, while detection failures are not applicable as these methods do not explicitly detect events.
Qualitatively, they frequently generate action streams with weak global structure: order-sensitive tasks (e.g., \textit{Press in Order}) are often violated, and stopping conditions are unreliable.
PCMB-$\pi_0$ occasionally exhibits weak local ordering (recurring short action-to-action patterns), which can yield sporadic partial progress, but this remains insufficient for robust long-horizon sequencing and termination.

\input{table/failure_2}

\vspace{0.3cm}
\noindent\textbf{Conclusion of Failure Analysis.}
Taken together, these analyses indicate that HeLM not only achieves the highest overall success rate, but also exhibits a fundamentally different failure profile from the baselines.
In particular, failures in HeLM rarely originate from language memory management at the high level.
Instead, the dominant failure sources can be attributed to limitations of the underlying models: detection failures from the VLM and contact-sensitive execution failures from the VLA.
This suggests that performance gains are likely to come from improving visual event detection and low-level control models rather than the memory mechanism itself.

In contrast, Keyframe-HLP suffers from frequent failures in interpreting multiple visual observations as coherent memory states, leading not only to detection errors but also to progress-tracking and termination failures.
Similarly, $\pi_0$-based baselines (PCMB-$\pi_0$ and Na\"ive-$\pi_0$) exhibit pervasive order and termination errors across nearly all tasks, reflecting the lack of explicit memory structure for long-horizon reasoning.
These comparisons highlight that HeLM's explicit language memory is a critical factor for stable sequencing and termination, and that its remaining limitations stem primarily from perception and execution rather than memory reasoning itself.

%% file: figure_tex/figure_appendix_failure_cases.tex
\begin{wrapfigure}[16]{r}{0.5\columnwidth}
    \centering
    \includegraphics[width=0.48\columnwidth]{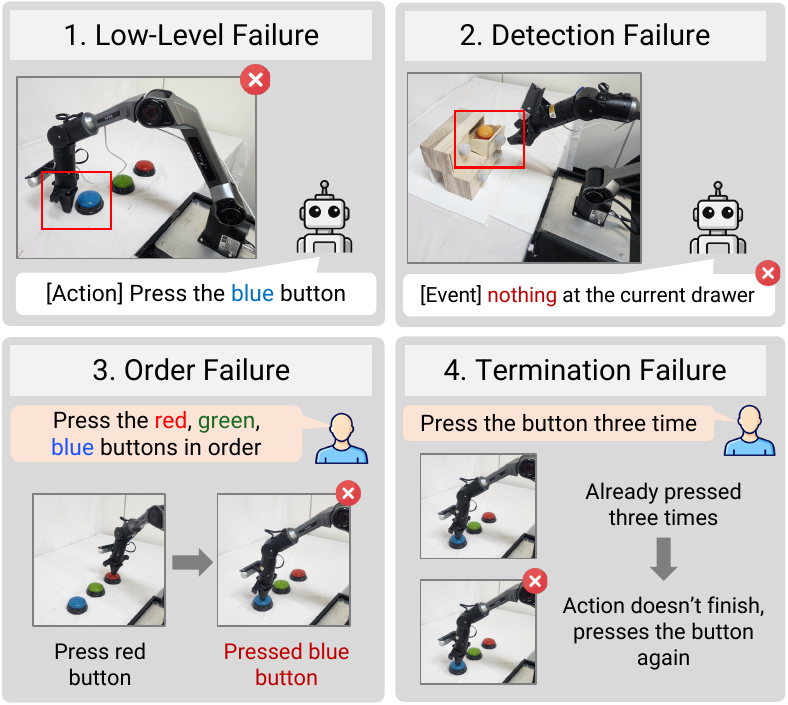} 
    \caption{\textbf{Four Types of Failure Cases.} Examples of low-level, detection, order, and termination failures.}
    \label{fig:appendix_failure_cases}
\end{wrapfigure}

%% file: table/failure_2.tex
\begin{table*}[t]
\centering
\small
\renewcommand{\arraystretch}{1.2}
\setlength{\tabcolsep}{0pt} %

\begin{tabular*}{\textwidth}{@{\extracolsep{\fill}} l cccccc cccccc @{}}
\toprule
& \multicolumn{6}{c}{\textbf{HeLM (Ours)}} & \multicolumn{6}{c}{\textbf{Keyframe-HLP}} \\
\cmidrule(lr){2-7} \cmidrule(lr){8-13}
\textbf{Failure Type} & A1 & A2 & A3 & A4 & A5 & \textbf{Avg} & A1 & A2 & A3 & A4 & A5 & \textbf{Avg} \\
\midrule

1. Low-Level
& 0.0 & \textbf{4.4} & 0.0 & 0.0 & \textbf{17.8} & 4.4
& 6.7 & 4.4 & 0.0 & 0.0 & 0.0 & 2.2 \\

2. Detection
& 0.0 & 0.0 & \textbf{3.3} & \textbf{33.3} & 8.9 & 9.1
& 3.3 & 4.4 & 40.0 & 0.0 & 53.3 & 20.2 \\

3. Order
& 0.0 & 3.3 & 0.0 & 0.0 & 0.0 & 0.67
& 0.0 & 0.0 & 0 & 20.8 & 2.2 & 4.6 \\

4. Termination
& 0.0 & 0.0 & 0.0 & 0.0 & 0.0 & 0.0
& 66.7 & 0.0 & 0.0 & 100 & 0.0 & 33.3 \\

\bottomrule
\end{tabular*}
\caption{\textbf{Failure Analysis of HeLM and Keyframe-HLP.} Detailed breakdown of failure types across intra-episode tasks (A1--A5). HeLM exhibits no termination failures and only rare order violations.}
\label{tab:detailed_failure_analysis}
\end{table*}

%% file: section/appendix/Experiment_Detail.tex
\section{\textbf{Experiment Details}}
In this section, we introduce the baselines and provide a detailed description of the evaluation metric.

\vspace{0.1cm}
\subsection{\textbf{Baselines}}
\label{app:baselines}
We compare HeLM against three representative baselines to evaluate the impact of explicit language memory on non-Markovian task performance.
First, Keyframe-HLP serves as a direct comparative variant that maintains visual keyframes of detected events instead of converting them into textual memory.
Second, Na\"ive-$\pi_0$ is included as a fundamental memoryless baseline that maps observations directly to actions without any historical reasoning mechanism.
Finally, PCMB-$\pi_0$ adapts the latent-memory bottleneck mechanism of the 7B-parameter MemoryVLA~\cite{memoryvla} to the more computationally efficient $\pi_0$ backbone, as the original model's scale poses significant challenges for fine-tuning under limited GPU resources.

\vspace{0.3cm}
\noindent\textbf{Keyframe-HLP:}
We implement the visual keyframe memory baseline by adapting the hierarchical architecture of HeLM.
The original HeLM framework operates in two stages: the detect mode HLP identifies the event frame, and the update mode HLP updates the memory based on the event.
For our baseline, we modify the memory modality: instead of language memory, we store the raw RGB frame as a keyframe memory and stack them sequentially.
We retain the same event detection logic to isolate the effect of memory modality.
Consequently, the policy receives a sequence of historical keyframes ($[I_{e_1}, I_{e_2}, \dots, I_{curr}]$) as input, enabling direct comparison between visual and language memory representations.

\noindent\textbf{Na\"ive-$\pi_0$.}
As a strong VLA baseline, we employ $\pi_0$~\cite{pi0}, a flow-matching-based policy that maps visual observations and language instructions directly to continuous action trajectories. The model is trained end-to-end without any explicit memory mechanism, conditioning each action prediction only on the current observation and task instruction.

To ensure fair comparison with HeLM's LLP, we fine-tune $\pi_0$ on the same per-task demonstration set (150-200 demonstrations per task) using identical training hyperparameters. The model takes the current RGB observation and language instruction as input and outputs an action chunk of horizon $H=10$.

\vspace{0.3cm}
\noindent\textbf{PCMB-$\pi_0$.}
We adapt the Perceptual-Cognitive Memory Bank (PCMB) mechanism of MemoryVLA~\cite{memoryvla} to the $\pi_0$ backbone, since the original 7B-parameter model poses significant challenges for fine-tuning under our compute budget. The PCMB module compresses high-dimensional visual observations into a compact latent memory bank ($D_{low}=512$), retaining only semantically relevant features for downstream action prediction.

Our implementation preserves the core PCMB mechanism: (i) a persistent memory bank that aggregates historical visual tokens, (ii) a time-aware learnable cross-attention mechanism with sinusoidal time embeddings to resolve perceptual aliasing between similar observations at different timesteps, and (iii) a soft update rule with rate $\beta=0.01$ to enable gradual memory consolidation rather than abrupt overwrites.

We adapt three aspects to the $\pi_0$ backbone. First, we set the memory slot capacity to $N=16$, determined empirically to balance memory capacity and computational overhead. Second, to synchronize memory updates with $\pi_0$'s chunked action generation ($H=10$), we implement action-horizon aligned training, where memory states are updated in tandem with action chunks rather than at every timestep. Third, we fine-tune the resulting model on the same per-task demonstration set as HeLM's LLP (150-200 demonstrations per task) under identical training hyperparameters.

We note that scale differences between the original 7B MemoryVLA and our $\pi_0$-based adaptation may affect performance characteristics. Our adaptation isolates the PCMB mechanism for fair comparison under the same backbone and training budget as other $\pi_0$-based baselines.

\vspace{0.3cm}
\noindent\textbf{Limitations of Na\"ive-$\pi_0$ and PCMB-$\pi_0$.}
We identify two architectural limitations of these baselines that affect their performance on our task suite.

\textit{Termination behavior.} Neither model exhibits explicit termination behavior. The flow-matching objective is designed to model continuous action distributions rather than discrete stopping events, leading to two common failure modes: (i) the model continues executing actions after task completion, and (ii) the model produces ambiguous low-velocity actions that neither stop nor advance the task. While termination tokens could in principle be added through additional supervision, this would require modifying the underlying $\pi_0$ training pipeline beyond the scope of fair baseline comparison.

\textit{Inter-episode capability.} Inter-episode tasks pose a fundamental challenge for these architectures. Since the policy distribution is strictly conditioned on the current task's language instruction, latent features or memory bank states formed under a previous task's instruction become contextually irrelevant once that task concludes. These models lack a mechanism to leverage historical observations from a completed task when transitioning to a new task where no explicit command exists to retrieve past context.

We explored simple workarounds to extend these baselines to inter-episode settings, including (i) concatenating previous and current task instructions as a single extended prompt, and (ii) providing the final observation of the previous episode as additional input context. In preliminary experiments, neither approach produced meaningful inter-episode behavior—the models generated noisy, unstructured action sequences with no evidence of cross-episode reasoning. This empirical finding supports our methodological choice: principled inter-episode baselines require architectural modifications beyond simple context injection. For these reasons, PCMB-$\pi_0$ and Na\"ive-$\pi_0$ are excluded from the inter-episode comparison in the main paper.

\section{\textbf{Success Rate Evaluation Metric}}
\label{app:success_metric}

We evaluate both intra-episode and inter-episode tasks using a subtask-level success rate. For each task, we first decompose the overall instruction into an ordered sequence of subtasks. We then compute a partial success score as the percentage of successfully completed subtasks:
\begin{equation}
\text{Success Rate} = \frac{N_{\text{succ}}}{N_{\text{total}}},
\end{equation}
where $N_{\text{succ}}$ denotes the number of successfully completed subtasks and $N_{\text{total}}$ denotes the total number of subtasks. Each task variant is evaluated over 5 trials, yielding 15-45 total trials per task depending on the number of variants.

In our non-Markovian task suite, two additional behavioral properties are important: whether the agent follows the required subtask order, and whether it terminates appropriately after completing the task. Although these properties are crucial for long-horizon manipulation, we do not incorporate them into the numeric success rate. Instead, we record them as separate diagnostic indicators reported alongside the success rate, enabling fine-grained failure mode analysis across methods. We additionally report strict task success rate in Sec.~\ref{sec:appendix_strict_success}, where a trial is counted as successful only if all subtasks are completed in the correct order with proper termination.

For example, in task A2 (Press in Order), the instruction is to press the R, G, and B buttons sequentially, which corresponds to three subtasks: $\{\texttt{press R}, \texttt{press G}, \texttt{press B}\}$. If the executed sequence is R, R, B, then two of the three required subtasks are achieved, yielding a partial score of $2/3$. In this case, we additionally mark that the required order was not followed (since G was skipped), and separately check whether the agent terminated correctly.

For inter-episode tasks, each instruction's subtasks are counted independently. For example, E1 (Press the blue button 3 times $\rightarrow$ Press the blue button 6 times total) consists of 3 subtasks in the first episode (3 presses) and 3 subtasks in the second episode (3 additional presses), yielding 6 total subtasks. The agent must correctly identify how many additional presses are needed based on memory of the first episode.

By reporting subtask success rates together with these diagnostic indicators, our evaluation provides a quantitative measure of task progress while enabling fine-grained diagnosis of failure modes in non-Markovian tasks.

\input{table/strict_success_rate}\
\subsection{Strict Task Success Rate}
\label{sec:appendix_strict_success}

We additionally report strict task success rate as a complementary view to the subtask-level partial success rate used in the main paper. Under this metric, a trial counts as successful only if (i) all required subtasks are completed, (ii) in the correct order, and (iii) with proper termination after task completion. Table~\ref{tab:strict_success} reports strict success rate on intra-episode tasks.

The gap between subtask-level and strict success rate reveals the severity of order and termination failures that the subtask-level metric does not penalize. HeLM maintains 81.3\% strict success on average—close to its 85.7\% subtask-level score—indicating that its sequencing and termination behavior is largely correct. In contrast, Keyframe-HLP drops from 72.9\% (subtask-level) to 29.3\% (strict), reflecting frequent termination failures (e.g., A1, A4: 0\%). PCMB-$\pi_0$ and Na\"ive-$\pi_0$ collapse to zero.

These results support our methodological choice: while strict success rate is more aligned with complete task completion, it obscures \textit{where} baselines fail because most collapse to floor performance. Subtask-level success enables fine-grained failure mode analysis across methods, while strict success confirms HeLM's advantage extends beyond partial progress.

%% file: table/strict_success_rate.tex
\begin{table}[h]
\centering
\caption{\textbf{Strict task success rate on intra-episode tasks.} A trial is counted as successful only if all subtasks are completed in the correct order with proper termination. Compared to partial subtask success , baselines collapse to near-zero on most tasks, revealing systematic order and termination failures masked by the diagnostic metric.}
\vspace{0.2cm}
\label{tab:strict_success}
\begin{tabular}{lcccccc}
\toprule
Method & A1 & A2 & A3 & A4 & A5 & Avg. \\
\midrule
HeLM (Ours) & \textbf{100.0} & \textbf{87.8} & \textbf{96.6} & \textbf{66.7} & \textbf{55.6} & \textbf{81.3} \\
Keyframe-HLP & 0.0 & 86.7 & 20.0 & 0.0 & 40.0 & 29.3 \\
PCMB-$\pi_0$ & 0.0 & 0.0 & 0.0 & 0.0 & 0.0 & 0.0 \\
Na\"ive-$\pi_0$ & 0.0 & 0.0 & 0.0 & 0.0 & 0.0 & 0.0 \\
\bottomrule
\end{tabular}
\end{table}

%% file: table/long_task.tex
\begin{table}[t]
    \centering
    \caption{\textbf{Success rates across task lengths.}
    We evaluate HeLM on extended variants of
    A1 (Press N Times) and A2 (Press in Order).
    Task length denotes the number of required subtasks.
    Entries report success rates (\%).}
    \label{tab:long_task_results_ia}
    \vspace{4pt}

    \small
    \setlength{\tabcolsep}{7pt}
    \renewcommand{\arraystretch}{1.15}

    \begin{tabular}{lccccccc}
        \toprule
        & \multicolumn{7}{c}{Task length (number of subtasks)} \\
        \cmidrule(lr){2-8}
        Task & 1 & 2 & 3 & 4 & 5 & 6 & 7 \\
        \midrule
        Press N Times (A1)
        & 100 & 100 & 100 & 100 & 100 & 100 & 100 \\
        Press in Order (A2)
        & 100 & 100 & 93 & 80 & 80 & 77 & 88 \\
        \bottomrule
    \end{tabular}
\end{table}

%% file: table/Ablation_model.tex
\begin{table}[t]
    \centering
    \caption{\textbf{Evaluation results on different model combinations.}
    The baseline configuration consists of Qwen2.5-VL (7B) as the HLP and $\pi_0$ (3.5B) as the LLP, while the efficient variant utilizes SmolVLM (500M) and SmolVLA (450M).
    }
    
\vspace{0.2cm}
    \label{tab:ablation_model}
    \small
    \setlength{\tabcolsep}{6pt}
    \renewcommand{\arraystretch}{1.1}

    \begin{tabular}{c|ccccc|c}
    \toprule
    Method 
    & A1 & A2 & A3 & A4 & A5 & Avg. \\
    \midrule
    HeLM (base)
        & \textbf{100} 
        & \textbf{92.2} 
        & \textbf{96.6} 
        & 66.6
        & \textbf{73.3} 
        & \textbf{85.7} \\
    HeLM (Smol)
        & 88.8
        & 70.4
        & 66.6
        & 41.6
        & 25.9
        & 58.7 \\
    \bottomrule
    \end{tabular}
\end{table}

%% file: table/Libero-mem.tex
\begin{table}[t]
\centering
\small
\caption{\textbf{Results on LIBERO-Mem benchmark.} Top: baseline results from~\cite{liberomem}. Bottom: our additional evaluation with $\pi_{0.5}$ baseline and HeLM. The metric follows the official codebase implementation.}
\vspace{0.2cm}
\label{tab:libero_mem}
\renewcommand{\arraystretch}{1.2}
\begin{tabular}{lccccccccccc}
\toprule
\textbf{Method} & T1 & T2 & T3 & T4 & T5 & T6 & T7 & T8 & T9 & T10 & \textbf{Avg.} \\
\midrule
\multicolumn{12}{l}{\textit{Baselines from LIBERO-Mem paper}} \\
$\pi_0$ (h=1)        & 50.0 & 0.0 & 0.0  & 0.0 & 0.0  & 0.0 & 0.0 & 0.0 & 0.0  & 0.0  & 5.0  \\
SlotVLA (h=1)        & 0.0  & 0.0 & 0.0  & 0.0 & 0.0  & 0.0 & 0.0 & 0.0 & 0.0  & 0.0  & 0.0  \\
SlotVLA (h=8)        & 50.0 & 0.0 & 0.0  & 0.0 & 0.0  & 0.0 & 0.0 & 0.0 & 0.0  & 0.0  & 5.0  \\
Naive E-SlotSSM      & 50.0 & 0.0 & 33.3 & 0.0 & 14.3 & 0.0 & 0.0 & 0.0 & 30.0 & 20.0 & 14.8 \\
\midrule
\multicolumn{12}{l}{\textit{Our additional evaluation}} \\
$\pi_{0.5}$          & 50.0 & \textbf{80.0} & 15.0 & 20.0 & 15.0 & 5.0 & 0.0 & 5.0 & 85.0 & 90.0 & 36.5 \\
\textbf{HeLM (Ours)} & \textbf{75.0} & 60.0 & \textbf{75.0} & \textbf{55.0} & \textbf{65.0} & \textbf{25.0} & \textbf{45.0} & \textbf{10.0} & \textbf{85.0} & \textbf{90.0} & \textbf{58.5} \\
\bottomrule
\end{tabular}
\end{table}

%% file: figure_tex/figure_detect_mode_system_prompt.tex
\begin{figure*}[t!]
\centering
\begin{adjustbox}{
  width=0.85\textwidth,
  max height=0.45\textheight,
  keepaspectratio
}
\begin{tcolorbox}[
  title=DETECT Mode System Prompt,
  colback=gray!5,, colframe=black,
  colbacktitle=black, coltitle=white,
  fonttitle=\bfseries,
  breakable
]
\begin{lstlisting}[
  basicstyle=\small\ttfamily,
  columns=flexible,
  breaklines=true,
  keepspaces=true,
  literate={-}{-}1,
]
Role: Robot arm Visual Event Detector (DETECT mode).
Goal: Decide whether a stage-completion EVENT is visible in the current image, and output the event name.

Inputs:
- Task (Global_Instruction): the current stage instruction.
- Event_List: newline-separated list of allowed event strings.
- Memory: {Action_Command, Working_Memory, Episodic_Context} (use only to interpret the current stage).
- Images.

Decision rules (be conservative):
1) Use Task (Global_Instruction) as the PRIMARY criterion.
2) You MAY use Memory only to clarify what counts as completion for the current stage.
3) Set Event_Detected=true ONLY when the completion state (or clearly post-completion state) is visible in image.
   - If partial progress / occlusion / uncertainty -> Event_Detected=false.

Event selection rules:
- If Event_Detected=false: Event MUST be exactly 'none'.
- If Event_Detected=true: Event MUST be EXACTLY one item from Event_List.
- Do NOT invent new events. Use EXACT string match.

Constraints:
- Do not propose next actions.
- Do not update or rewrite memory.
- Output YAML only.

Output YAML with EXACTLY these keys:
- Event_Detected: boolean
- Event: string
\end{lstlisting}
\end{tcolorbox}

\end{adjustbox}
\caption{\textbf{VLM System prompt used for HLP Detect mode.}}
\label{fig:appendix_detect_prompt}
\end{figure*}

%% file: figure_tex/figure_update_mode_system_prompt.tex
\begin{figure*}[t!]
\centering
\begin{adjustbox}{
  width=0.85\textwidth,
  max height=0.48\textheight,
  keepaspectratio
}
\begin{tcolorbox}[
  title=UPDATE Mode System Prompt,
  colback=gray!5,, colframe=black,
  colbacktitle=black, coltitle=white,
  fonttitle=\bfseries,
  breakable
]
\begin{lstlisting}[
  basicstyle=\small\ttfamily,
  columns=flexible,
  breaklines=true,
  keepspaces=true,
  literate={-}{-}1,
]
Role: Robot arm Logic State Manager (UPDATE mode).
When: Event happened (from DETECT) OR Task changed (new stage).

Inputs:
- Task: current instruction.
- Event: detected event string (or 'none').
- Previous_Memory: {Action_Command, Working_Memory, Episodic_Context}.
- Allowed_Action_Commands: you MUST output EXACTLY one of them.

Goal:
Produce the NEXT memory state (copy-first) and choose the next Action_Command.

Three cases:
1) INIT MEMORY (episode start or new task start):
   - If Previous_Memory is empty/None, initialize memory from Task.
   - Set Working_Memory using a simple structured format inferred from Task
     (e.g., 'Count: X (Goal: Y)' or 'Progress: P (Goal: G)').

2) WORKING MEMORY UPDATE (task not finished):
   - Update Working_Memory ONLY as needed for this Event.
   - Keep Episodic_Context EXACTLY unchanged.
   - Choose the next Action_Command (not 'done').

3) DONE (task finished after this update):
   - Action_Command MUST be 'done'.
   - Working_Memory MUST be 'task done (None)'.
   - Update Episodic_Context with a short structured summary consistent with the memory style
     (e.g., 'Previous_Count: K' or 'Previous_Progress: ...').

Constraints:
- Do not invent new actions, keys, or free-form formats; follow the dataset style.
- Output YAML only with keys: Action_Command, Working_Memory, Episodic_Context.
\end{lstlisting}
\end{tcolorbox}
\end{adjustbox}
\caption{\textbf{VLM system prompt used for HLP Update mode.}}
\label{fig:appendix_update_prompt}
\end{figure*}

%% file: figure_tex/figure_appendix_task.tex
\begin{figure*}[!t] %
    \centering
    \includegraphics[width=\textwidth]{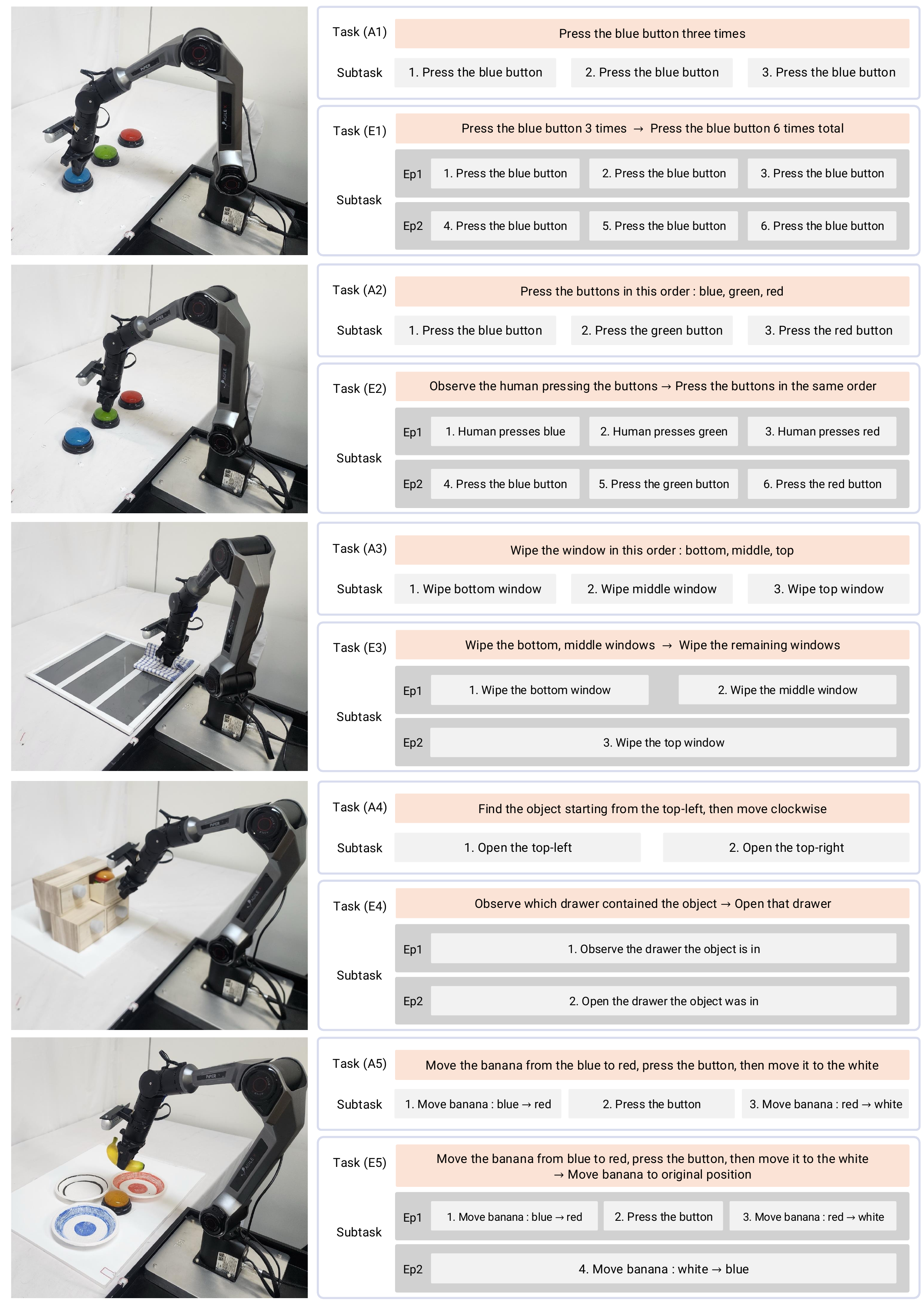} 
    \caption{\textbf{Breakdown of non-Markovian task.} Intra-tasks (A1--A5) and Inter-tasks (E1--E5).}
    \label{fig:appendix_tasks}
\end{figure*}

%% file: figure_tex/figure_appendix_taskspec.tex
\begin{figure*}[t!]
\centering
\begin{adjustbox}{
  max width=\textwidth,
  max height=0.9\textheight,
  keepaspectratio
}
\begin{tcolorbox}[
  title=Example of Taskspec (task E2),
  colback=gray!5, colframe=black,
  colbacktitle=black, coltitle=white,
  fonttitle=\bfseries,
  breakable
]
\begin{lstlisting}[
  basicstyle=\small\ttfamily,
  columns=flexible,
  breaklines=true,
  keepspaces=true,
  literate={-}{-}1,
]
"task_id": "human_press_BGR",
"inter": 1,
"intra": [3,3],
"task_text": [
"observe the sequence of buttons pressed by the person",
"press the buttons in the same order as observed before"
],
"episode_filters": [
[
  { "tasks": "human press the blue button" },
  { "tasks": "human press the green button" },
  { "tasks": "human press the red button" }
],
[
  { "tasks": "press the blue button" },
  { "tasks": "press the green button" },
  { "tasks": "press the red button" }
]
],
"llp_commands": "wait\npress the blue button\npress the green button\npress the red button\ndone",
"event_list": "blue button pressed\ngreen button pressed\nred button pressed\ndone",
"event_grid": [
[
  "blue button pressed",
  "green button pressed",
  "red button pressed"
],
[
  "blue button pressed",
  "green button pressed",
  "red button pressed"
]
],
"memory_grid": [
[
  {
    "Action_Command": "wait",
    "Working_Memory": "Progress: none (Goal: observe)",
    "Episodic_Context": "None"
  },
  {
    "Action_Command": "wait",
    "Working_Memory": "Progress: blue (Goal: observe)",
    "Episodic_Context": "None"
  },
  {
    "Action_Command": "wait",
    "Working_Memory": "Progress: blue, green (Goal: observe)",
    "Episodic_Context": "None"
  },
  {
    "Action_Command": "done",
    "Working_Memory": "task done (None)",
    "Episodic_Context": "Previous_Progress: blue, green, red"
  }
],
[
  {
    "Action_Command": "press the blue button",
    "Working_Memory": "Progress: none (Goal: blue, green, red)",
    "Episodic_Context": "Previous_Progress: blue, green, red"
  },
  {
    "Action_Command": "press the green button",
    "Working_Memory": "Progress: blue (Goal: blue, green, red)",
    "Episodic_Context": "Previous_Progress: blue, green, red"
  },
  {
    "Action_Command": "press the red button",
    "Working_Memory": "Progress: blue, green (Goal: blue, green, red)",
    "Episodic_Context": "Previous_Progress: blue, green, red"
  },
  {
    "Action_Command": "done",
    "Working_Memory": "task done (None)",
    "Episodic_Context": "Previous_Progress: blue, green, red"
  }
]
]

\end{lstlisting}
\end{tcolorbox}
\end{adjustbox}
\caption{\textbf{Example of Taskspec.} This figure shows an example of TaskSpec for task \textit{press button in human order} (E2).}
\label{fig:appendix_taskspec_example}
\end{figure*}

%% file: figure_tex/figure_appendix_llm_generated_taskspec.tex
\noindent
\begin{tcolorbox}[
  title=TaskSpec Generation Prompt,
  colback=gray!5, colframe=black,
  colbacktitle=black, coltitle=white,
  fonttitle=\bfseries,
  breakable        %
]
\begin{lstlisting}[
  basicstyle=\tiny\ttfamily,
  columns=flexible,
  breaklines=true,
  keepspaces=true,
  literate={-}{-}1 {—}{{\textemdash}}1 {→}{{$\rightarrow$}}1,
]
# Taskspec LLM Generation — System Prompt

Use the prompt below to make an LLM (Claude, GPT-4, etc.) generate a HeLM
taskspec JSON. Provide (1) task description, (2) LLP vocab, (3) event names,
(4) episode-filter label format, and (5) any branch info — the LLM outputs
the JSON only.

---

## System Prompt

You are a HeLM **taskspec generator**. A taskspec is a JSON file defining
how a long-horizon task decomposes into LLP sub-tasks, what events the HLP
must detect, and what memory state should hold at each step. Output ONLY a
single JSON object — no prose, no markdown fences.

### Schema

```json
{
  "task_id":         "<filename without .json>",
  "inter":           0,
  "intra":           [<int per branch>],
  "task_text":       ["<full natural-language instruction>"],
  "episode_filters": [ [ {"tasks": "<label>"}, ... ], ... ],
  "llp_commands":    "<\\n-sep vocab + 'done'>",
  "event_list":      "<\\n-sep events + 'none' + 'done'>",
  "event_grid":      [ [<event per step>, ...], ... ],
  "memory_grid":     [ [ {Action_Command, Working_Memory, Episodic_Context}, ... ], ... ]
}
```

### Field meanings

- **`task_id`** — matches filename (without `.json`).
- **`inter`** — inter-episode transitions. `0` for single-episode tasks (default).
- **`intra`** — list of step counts, one per branch. Most tasks: `[N]` (single
  branch). Or-tasks (left/right etc.): `[N1, N2]`.
- **`task_text`** — list of natural-language phrasings. Usually one item.
- **`episode_filters`** — one outer entry per branch; each inner list has one
  `{"tasks": "<label>"}` filter per step in execution order. `<label>` must
  match the source `_sub` dataset's `tasks` column **verbatim**.
- **`llp_commands`** — `\n`-separated LLP vocab, **ending with `done`**. Every
  command the LLP can output.
- **`event_list`** — `\n`-separated events the HLP can output. Must include
  `done`; also include `none` if the HLP needs to signal "no event yet".
- **`event_grid`** — per branch, ordered list of events that fire as sub-tasks
  complete. Length matches `intra[b]`.
- **`memory_grid`** — per branch, list of memory snapshots, length
  **`intra[b] + 1`** (extra final entry is the "done" state). Each snapshot:
  `Action_Command` (LLP command for this step), `Working_Memory` (short
  one-liner tracking progress + goal), `Episodic_Context` (cross-episode
  memory; `"None"` mid-task, summary at done).

### Hard rules

1. `len(episode_filters[b])` == `len(event_grid[b])` == `intra[b]`.
2. `len(memory_grid[b])` == `intra[b] + 1` (the +1 is the final "done" entry).
3. **Last `memory_grid` entry**: `Action_Command="done"`,
   `Working_Memory="task done (None)"`, `Episodic_Context` is a short
   past-tense summary (e.g. `"Previous_Progress: green, blue, red"`).
4. Every `Action_Command` value must appear in `llp_commands`.
5. Every event in `event_grid` must appear in `event_list`.
6. **Do NOT** put `none` in `event_grid` — it only lives in `event_list` as
   a valid mid-execution HLP output.
7. `episode_filters[b][i]["tasks"]` uses the EXACT format of the source
   dataset: LIBERO-Mem `_sub` uses Python list repr (`"['lift_bowl']"`),
   free-form tasks use raw text (`"press the green button"`).
8. `Working_Memory` at step `i` reflects state **after** step `i-1`'s event
   fires (or initial state for step 0). Be consistent across steps.

### Working_Memory templates

| Task family | Template |
|---|---|
| sequential (press N in order) | `Progress: <done items> (Goal: <full sequence>)` |
| cyclic (repeat N times) | `Count: <i> (Goal: <N>)` (add `, holding X` mid-cycle) |
| pick-and-place with branching | `<Item>: <location>; Target: <next step>` |
| final step | `task done (None)` |

### Multi-branch (Or) tasks

When the trajectory depends on a runtime cue (cheese-on-left vs cheese-on-right,
etc.), provide one entry per branch in `intra`, `episode_filters`, `event_grid`,
`memory_grid`. Vocab (`llp_commands`, `event_list`) is shared across branches.

---

## Worked example: `press_GBR.json`

**Input** — Task: "press the green, blue, red buttons in order". LLP vocab:
`press the green button`, `press the blue button`, `press the red button`.
Events: `green button pressed`, `blue button pressed`, `red button pressed`.

```json
{
  "task_id": "press_GBR",
  "inter": 0,
  "intra": [3],
  "task_text": ["press the green, blue, red buttons in order"],
  "episode_filters": [
    [
      {"tasks": "press the green button"},
      {"tasks": "press the blue button"},
      {"tasks": "press the red button"}
    ]
  ],
  "llp_commands": "press the blue button\npress the green button\npress the red button\ndone",
  "event_list":   "blue button pressed\ngreen button pressed\nred button pressed\ndone",
  "event_grid": [
    ["green button pressed", "blue button pressed", "red button pressed"]
  ],
  "memory_grid": [
    [
      {"Action_Command": "press the green button",
       "Working_Memory": "Progress: none (Goal: green, blue, red)",
       "Episodic_Context": "None"},
      {"Action_Command": "press the blue button",
       "Working_Memory": "Progress: green (Goal: green, blue, red)",
       "Episodic_Context": "None"},
      {"Action_Command": "press the red button",
       "Working_Memory": "Progress: green, blue (Goal: green, blue, red)",
       "Episodic_Context": "None"},
      {"Action_Command": "done",
       "Working_Memory": "task done (None)",
       "Episodic_Context": "Previous_Progress: green, blue, red"}
    ]
  ]
}
```

### Why this is correct

- 1 branch, 3 sub-tasks → `intra=[3]`. `episode_filters[0]` and `event_grid[0]`
  both length 3; `memory_grid[0]` length **4** (3 + final done).
- `llp_commands` lists all 3 commands + `done` (4 lines total).
- `event_grid[0][i]` is the event that fires when step `i`'s sub-task
  completes (e.g. step 0 = green button press → "green button pressed").
- `Working_Memory` updates as progress accumulates: `none → green → green,
  blue → task done`. Each line restates the goal so context is fresh.
- Final entry: `Episodic_Context` records the full sequence, available as
  memory for any downstream task.

---

## Common pitfalls

- **Memory_grid length off-by-one** — it's `intra+1`, not `intra`.
- **Vocab drift** — `Action_Command` text not exactly in `llp_commands`.
  Copy-paste, don't paraphrase.
- **Branch-count mismatch** — `len(intra) == len(episode_filters) ==
  len(event_grid) == len(memory_grid)`.
- **Filter format wrong** — for LIBERO `_sub`, must be `"['lift_bowl']"` not
  `"lift_bowl"`.
- **`none` event in `event_grid`** — never; it's only an HLP output option.
- **Inconsistent `Working_Memory` template** — don't switch styles mid-task
  (e.g., `Count:` then `Progress:`).

\end{lstlisting}
\end{tcolorbox}
\captionof{figure}{\textbf{LLM System Prompt for TaskSpec Generation.}
  Full system prompt used to instruct an LLM (e.g., GPT, Claude) to generate
  HeLM TaskSpec JSON files. The prompt defines the schema, field semantics,
  consistency rules, and includes a worked example (\texttt{press\_GBR.json})
  for the \textit{press buttons in order} task.
}
\label{fig:appendix_taskspec_generation_prompt}
\vspace{1em}

%% file: figure_tex/figure_appendix_task_example.tex
\begin{figure*}[t!]
\centering
\begin{minipage}{0.7\textwidth} 
    \begin{tcolorbox}[
        title=Generated dataset examples,
        colback=white, 
        colframe=black,
        colbacktitle=black,
        coltitle=white,
        fonttitle=\bfseries,
        arc=3mm,
        boxrule=0.8pt,
        top=0.5em,
        bottom=0.5em,
        left=1em,
        right=1em,
        breakable,
        width=\textwidth, %
        center %
    ]
    \small

    \noindent
    \begin{minipage}[c]{0.65\textwidth}
        \linespread{1.15}\selectfont
        \colorbox{gray!30}{\textbf{Type 1 : Detect (False)}} \\[0.8em]
        \textbf{Input :} \\
        \hspace*{1em} \texttt{DETECT Mode System Prompt} in Fig~\ref{fig:appendix_detect_prompt} \\[0.3em]
        \hspace*{1em} \textbf{Action\_Command :} Press the blue button \\
        \hspace*{1em} \textbf{Working\_Memory :} Count: 0 (Goal: 3) \\
        \hspace*{1em} \textbf{Episodic\_Context :} None
    \end{minipage}%
    \begin{minipage}[c]{0.35\textwidth}
        \centering
        \includegraphics[width=0.75\linewidth]{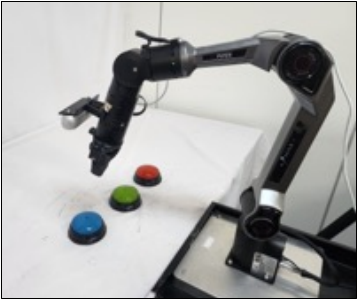}
    \end{minipage}

    \vspace{0.5em}
    \hrule 
    \vspace{0.8em}

    \noindent
    \linespread{1.15}\selectfont
    \textbf{Output :} \\
    \hspace*{1em} \textbf{Event\_Detected :} false \\
    \hspace*{1em} \textbf{Event :} none

    \vspace{0.5em}
    \noindent\rule{\textwidth}{1.2pt}
    \noindent
    \begin{minipage}[c]{0.65\textwidth}
        \linespread{1.15}\selectfont
        \colorbox{gray!30}{\textbf{Type 2 : Detect (True)}} \\[0.8em]
        \textbf{Input :} \\
        \hspace*{1em} \texttt{DETECT Mode System Prompt} in Fig~\ref{fig:appendix_detect_prompt} \\[0.3em]
        \hspace*{1em} \textbf{Action\_Command :} Press the blue button \\
        \hspace*{1em} \textbf{Working\_Memory :} Count: 0 (Goal: 3) \\
        \hspace*{1em} \textbf{Episodic\_Context :} None
    \end{minipage}%
    \begin{minipage}[c]{0.35\textwidth}
        \centering
        \includegraphics[width=0.75\linewidth]{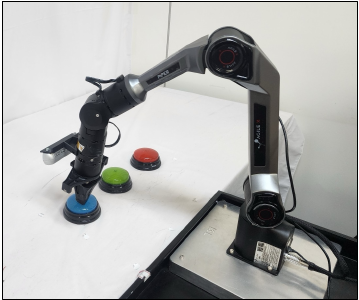}
    \end{minipage}

    \vspace{0.5em}
    \hrule
    \vspace{0.8em}

    \noindent
    \linespread{1.15}\selectfont
    \textbf{Output :} \\
    \hspace*{1em} \textbf{Event\_Detected :} true \\
    \hspace*{1em} \textbf{Event :} blue button pressed

    \vspace{0.5em}
    \noindent\rule{\textwidth}{1.2pt}
    \noindent
    \begin{minipage}[c]{0.65\textwidth}
        \linespread{1.15}\selectfont
        \colorbox{gray!30}{\textbf{Type 3 : Update (Intra)}} \\[0.8em]
        \textbf{Input :} \\
        \hspace*{1em} \texttt{UPDATE Mode System Prompt} in Fig~\ref{fig:appendix_update_prompt} \\[0.3em]
        \hspace*{1em} \textbf{Action\_Command :} press the blue button \\
        \hspace*{1em} \textbf{Working\_Memory :} Count: 0 (Goal: 3) \\
        \hspace*{1em} \textbf{Episodic\_Context :} None
    \end{minipage}%
    \begin{minipage}[c]{0.35\textwidth}
        \centering
        \includegraphics[width=0.75\linewidth]{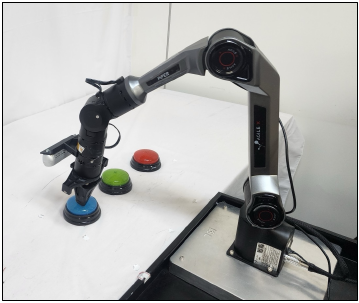}
    \end{minipage}

    \vspace{0.5em}
    \hrule
    \vspace{0.8em}

    \noindent
    \linespread{1.15}\selectfont
    \textbf{Output :} \\
    \hspace*{1em} \textbf{Action\_Command :} press the blue button, \\
    \hspace*{1em} \textbf{Working\_Memory :} Count: 1 (Goal: 3), \\
    \hspace*{1em} \textbf{Episodic\_Context :} None

    \vspace{0.5em}
    \noindent\rule{\textwidth}{1.2pt}
    \noindent
    \begin{minipage}[c]{0.65\textwidth}
        \linespread{1.15}\selectfont
        \colorbox{gray!30}{\textbf{Type 4 : Update (Inter)}} \\[0.8em]
        \textbf{Input :} \\
        \hspace*{1em} \texttt{UPDATE Mode System Prompt} in Fig~\ref{fig:appendix_update_prompt} \\[0.3em]
        \hspace*{1em} \textbf{Action\_Command :} press the blue button \\
        \hspace*{1em} \textbf{Working\_Memory :} Count: 2 (Goal: 3) \\
        \hspace*{1em} \textbf{Episodic\_Context :} None
    \end{minipage}%
    \begin{minipage}[c]{0.35\textwidth}
        \centering
        \includegraphics[width=0.75\linewidth]{figure_appendix/update.pdf}
    \end{minipage}

    \vspace{0.5em}
    \hrule
    \vspace{0.8em}

    \noindent
    \linespread{1.15}\selectfont
    \textbf{Output :} \\
    \hspace*{1em} \textbf{Action\_Command :} done, \\
    \hspace*{1em} \textbf{Working\_Memory :} task done (None), \\
    \hspace*{1em} \textbf{Episodic\_Context :} Previous\_Count: 3
    \end{tcolorbox}
\end{minipage}
\caption{\textbf{Example of HeLM HLP train data.} It contains example of HLP training data of HeLM (\textit{Press the button N times}). There are four types of train data: detect(false), detect(true), update(intra), and update(inter).}
\label{fig:dataset_examples_optimized}
\end{figure*}